\documentclass[final]{cvpr}
\usepackage{times}
\usepackage{epsfig}
\usepackage{graphicx}
\usepackage{amsmath}
\usepackage{amssymb}
\usepackage{subfigure}
\usepackage{multirow}
\newcommand{\vect}[1]{\mathbf{#1}}
\newcommand{\matr}[1]{\mathbf{#1}}

\usepackage[pagebackref=true,breaklinks=true,colorlinks,bookmarks=false]{hyperref}

\def\cvprPaperID{6913} 
\def\confYear{CVPR 2021}

\begin{document}


\title{3D AffordanceNet: A Benchmark for Visual Object Affordance Understanding}


\author{Shengheng Deng\textsuperscript{1,*},
Xun Xu\textsuperscript{2,*}, Chaozheng Wu\textsuperscript{1},
Ke Chen\textsuperscript{1,4} and Kui Jia\textsuperscript{1,3,4,\textdagger}\\
\textsuperscript{1}South China University of Technology,\quad \textsuperscript{2}I2R, A-STAR\\ \textsuperscript{3}Pazhou Laboratory, \quad \textsuperscript{4}Peng Cheng Laboratory\\
}

\maketitle
\let\thefootnote\relax\footnote{* indicates equal contribution.}

\let\thefootnote\relax\footnote{\textsuperscript{\textdagger}Correspondence to Kui Jia $<$kuijia@scut.edu.cn$>$.}

\begin{abstract}
The ability to understand the ways to interact with objects from visual cues, a.k.a. visual affordance, is essential to vision-guided robotic research. This involves categorizing, segmenting and reasoning of visual affordance. Relevant studies in 2D and 2.5D image domains have been made previously, however, a truly functional understanding of object affordance requires learning and prediction in the 3D physical domain, which is still absent in the community. In this work, we present a 3D AffordanceNet dataset, a benchmark of 23k shapes from 23 semantic object categories, annotated with 18 visual affordance categories. Based on this dataset, we provide three benchmarking tasks for evaluating visual affordance understanding, including full-shape, partial-view and rotation-invariant affordance estimations. Three state-of-the-art point cloud deep learning networks are evaluated on all tasks. In addition we also investigate a semi-supervised learning setup to explore the possibility to benefit from unlabeled data. Comprehensive results on our contributed dataset show the promise of visual affordance understanding as a valuable yet challenging benchmark.  
\end{abstract}

\section{Introduction}


\begin{figure}[t]
   \begin{center}
      \includegraphics[width=0.95\linewidth]{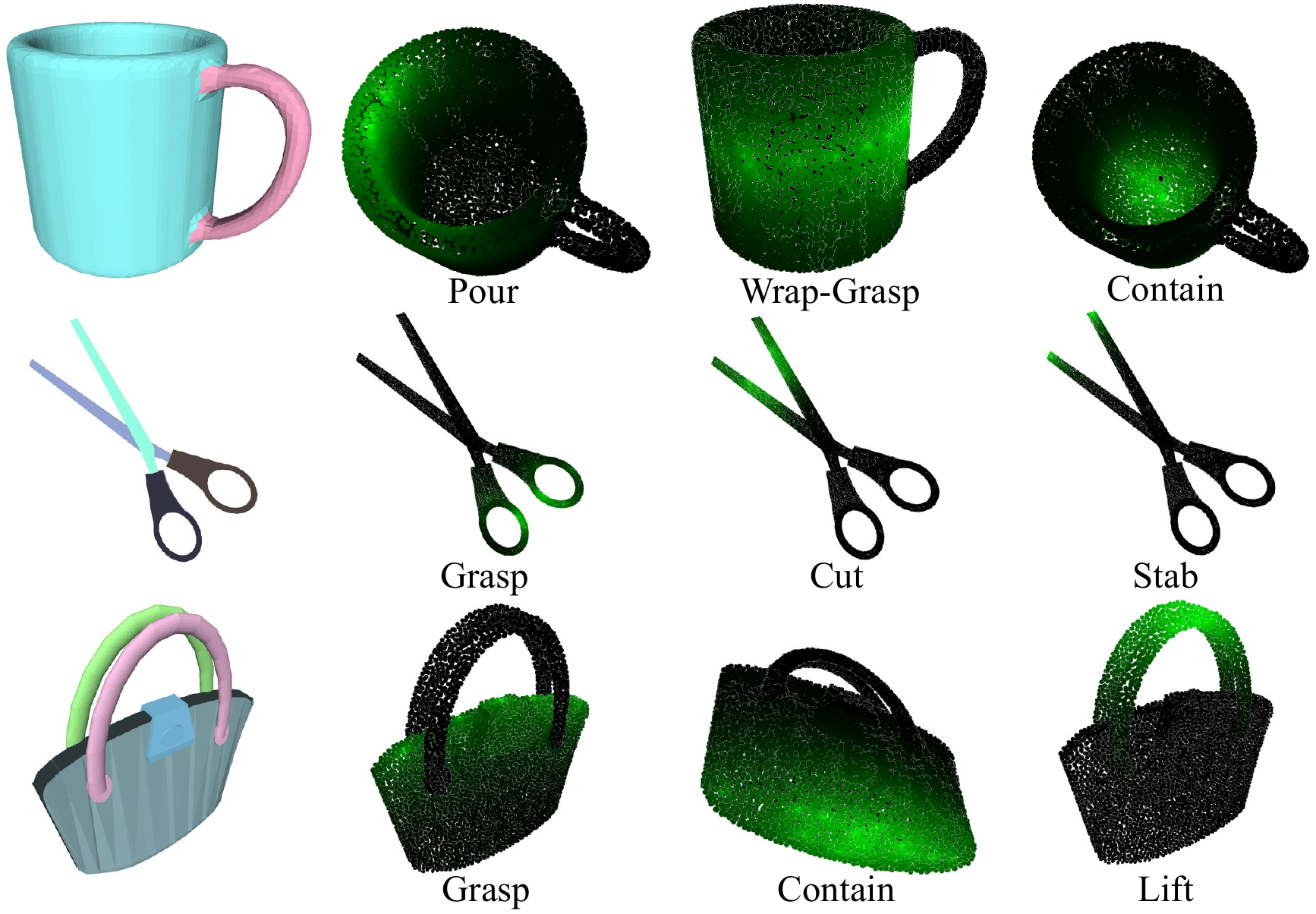}
   \end{center}
   \vspace{-0.4cm}
   \caption{The 3D AffordanceNet dataset. The mesh was first annotated with affordance keypoints. Then we densely sample points and obtain the ground truth data via label propagation.}
   \label{annotation image}
   \vspace{-0.6cm}
\end{figure}

The concept of affordance was first defined as what the environment  offers  the  animal,  introduced by \cite{gibson1979ecological}.
Affordance understanding is concerned with the interactions between human and environment. For instance, human can sit on the chair, grasp a cup or lift a bag. Being able to understand the affordance of objects is crucial for robots to operate in dynamic and complex environments \cite{hassanin2018visual}. Many applications are supported by affordance understanding including, anticipating and predicting future actions\cite{koppula2013learning,jain2016structural,koppula2015anticipating},
recognizing agent's activities\cite{qi2017predicting,earley1970efficient,vu2014predicting},
providing valid functionality of the objects\cite{grabner2011makes},
understanding social scene situations\cite{chuang2018learning}
and understanding the hidden values of the objects\cite{zhu2015understanding}. Tasks including affordance categorization, reasoning, semantic labeling, activity recognition, etc. are defined as specific instantiations of affordance understanding \cite{hassanin2018visual}. Among all these we find semantic labeling \cite{roy2016multi,zhu2015understanding} is of the most importance because the ability to localize the position of possible affordance is highly desired by robotic research. We refer semantic labeling as affordance estimation throughout this paper.


The most important and proper modality for affordance understanding is through visual sensors \cite{hassanin2018visual}. Visual affordance understanding has been extensively studied recently with computer vision techniques. Many algorithms are built upon deep neural networks \cite{nguyen2017object,do2018affordancenet,sawatzky2017weakly} thus require large labeled affordance dataset for benchmarking. Relevant datasets are developed for these purposes with data collected from 2D (RGB) sensors \cite{zhou2019semantic,song2015learning,roy2016multi} or 2.5D (RGBD) sensors \cite{myers2015affordance,nguyen2017object,sawatzky2017weakly}. Nevertheless, we believe that the affordance understanding requires learning in the 3D domain which conveys the geometric properties. For example, the affordance of grasp is highly correlated with vertical structure with small perimeter and sittable is correlated with flat surface. Unfortunately, such detailed geometry is not captured by the existing 2D datasets while the 2.5 ones \cite{nguyen2017object,sawatzky2017weakly} are often captured with small depth variation and do not carry enough geometric information.



To encourage research into visual affordance understanding in more realistic scenarios, a benchmark on real 3D dataset is highly desired. Therefore, we are inspired by PartNet\cite{mo2019partnet}, a recently proposed dataset containing the fine-grained part
hierarchy information of 3D shapes based on the large-scale 3D CAD model
dataset ShapeNet\cite{chang2015shapenet} and 3D Warehouse\footnote{https://3dwarehouse.sketchup.com}.
Although PartNet mentioned affordance as potential application, there is still no benchmark purposely established for affordance yet. More importantly, we discover, via user annotations, that the human perceived affordance often do not fully overlap with the individual parts specified in PartNet dataset. For example, In the first row of Fig.~\ref{annotation image}, the \textit{Pour}, \textit{Wrap-Grasp} and \textit{Contain} affordance from \textit{Mug} do not perfectly match any part indicated by the colored image on the 1st column. Therefore, we believe it is necessary to provide a new set of affordance labels on the PartNet dataset.

Creating 3D visual affordance benchmark is challenging due to the subjective definition. We take into account the affordance definitions from existing research on visual affordance learning in 2D and 2.5D domains \cite{hassanin2018visual} and select possible interactions that one can take with 3D shapes from PartNet. Finally, 18 types of affordance were formally defined over 23 semantic objects. Additional challenge associated with annotation on 3D model is the scalability issue. In order to provide highly quality annotation on such a large scale, we use label propagation method to propagate affordance sparsely labeled on individual points. Eventually, we obtain point-wise probabilistic score of affordance for each individual shape in PartNet. We name the new benchmark \textit{3D AffordanceNet} to reflect the focus on visual affordance on 3D point cloud data.

3D AffordanceNet enables benchmarking a diverse set of tasks, in particular, we put forward full-shape, partial-view and rotation-invariant affordance estimations. Three state-of-the-art point cloud deep learning networks are evaluated on all tasks. We also propose a semi-supervised affordance estimation method to take the advantage of large amount of unlabeled data for affordance estimation.



In summary, we make the following contributions:
\begin{itemize}
   \item We introduce 3D AffordanceNet, consisting of 56307 well-defined affordance
         information annotations for 22949 shapes covering 18 affordance classes and
         23 semantic object categories.
         To the best of our knowledge, this is the first \textit{large-scale} dataset with
         \textit{well-defined} probabilistic affordance score annotations;
   \item We propose three affordance learning tasks which are supported by 3D AfffodanceNet
         to demonstrate the value of annotated data: full-shape affordance estimation, partial-view affordance estimation and rotation-invariant affordance estimation.
   \item We benchmark three baseline methods for proposed affordance learning tasks and further propose a semi-supervised affordance estimation method to take advantage of unlabeled data for affordance estimation. 
\end{itemize}
\section{Related Work}

\begin{figure*}[t]
   \begin{center}
      \includegraphics[width=0.9\linewidth]{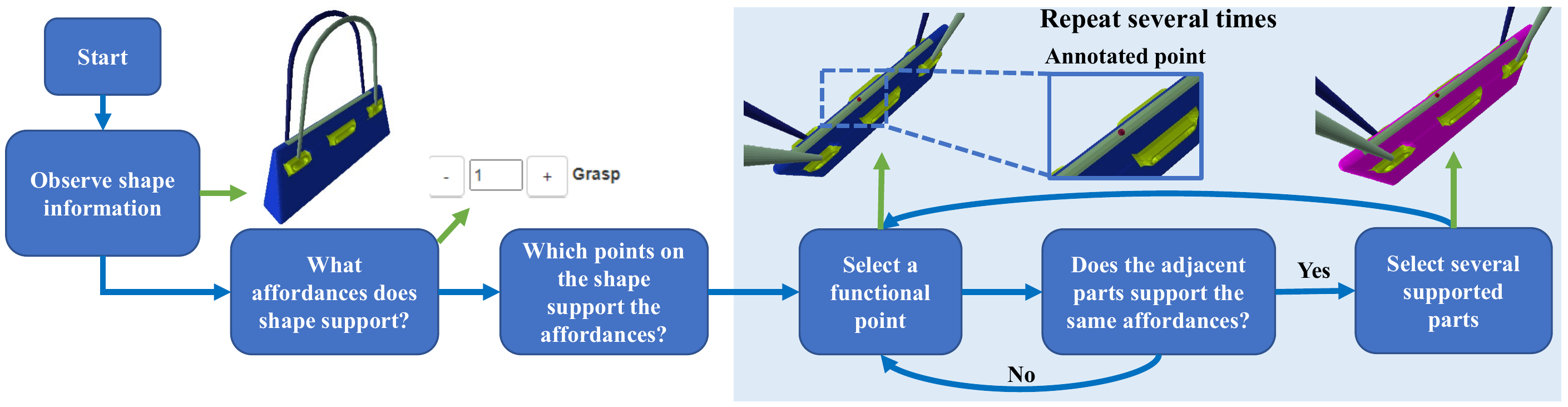}
   \end{center}
   \vspace{-0.4cm}
   \caption{The annotation workflow. The blue arrows indicate the annotation procedures,
      the green arrows refer to the corresponding 3D GUI actions. The annotators are first asked to
      determine the supported affordance classes and then select the functional points. The annotators
      need to confirm whether the adjacent parts support the same affordances.}
   \label{figure.interface}
   \vspace{-0.5cm}
\end{figure*}

Affordance refers to the possible action an agent could make to interact with the environment \cite{gibson1979ecological}. Examples include a cup can afford `\textit{pouring}', a bed is `\textit{sittable}' and `\textit{layable}', \etc. Affordance understanding is the core function in developing autonomous systems. In particular, the visual affordance understanding is the most promising way due to the rich information carried by visual sensors. We mainly review the recent development in visual affordance dataset and approaches, a detailed review can be found in \cite{hassanin2018visual}. Recent advances in visual affordance are mostly demonstrated on affordance recognition \cite{grabner2011makes,do2018affordancenet}, detection \cite{do2018affordancenet, myers2015affordance} and segmentation \cite{do2018affordancenet, myers2015affordance}. Beyond the low-level visual tasks, there is substantial attention on-the-rise paid to affordance reasoning, affordance-based activity recognition and social affordances. In this work we are interested in providing a benchmark for affordance segmentation, a.k.a. prediction, due to the clear definition and high demand in robotic applications. 

To benchmark visual affordance segmentation, UMD \cite{myers2015affordance}, CAD120 \cite{sawatzky2017weakly} and IIT-AFF \cite{nguyen2017object} are respectively developed recently. All datasets feature affordance segmentation on RGBD images covering from 10-20 objects, 6-9 affordance types and 3k-10k labelled images. In particular, IAF-IIT and CAD120 capture complex scenes while UMD mainly focuses on well-controlled scenes. Nevertheless, none of these datasets carry the rich geometric properties of objects that robotic application would expect and only a single view-point is present. As a result, these datasets no longer pose challenges to modern computer vision techniques.

With the easy access to 3D point cloud data, \eg collected from LiDAR and SFM, and potential application in robotics, atunomous driving, \etc., there is a recent surge in research towards 3D point cloud. ShapeNet \cite{chang2015shapenet} collected
3D CAD models from open-sourced 3D repositories,
with more than 3,000,000 models and 3,135 object categories. It was further developed by \cite{yi2016scalable} for shape part segmentation. Partially motivated by affordance understanding, the subsequent PartNet dataset \cite{mo2019partnet} was proposed with 26k objects, featuring fine-grained semantic segmentation task. Hierarchical segmentation was also addressed by a recursive part decomposition \cite{yu2019partnet}. Though affordance is briefly mentioned as the motivation for creating above 3D shape datasets, to the best of our knowledge, there is no existing dataset which explicitly addresses the task of visual affordance prediction. The only known attempt on 3D shape functionality understanding \cite{hu2016learning} is still limited to a few types of objects and did not make connection to the well-studied visual affordance understanding in 2D image domains. In contrast, we created a new benchmark for visual affordance estimation on 3D point cloud. The affordance types are selectively inherited from a summary of existing works and annotations are made on 3D point cloud data directly.

\section{Dataset Construction}




We present 3D AffordanceNet as a dataset for affordance estimation 3D point cloud. To construct this dataset, we first define a set of affordance types by referring to the existing visual affordance works \cite{hassanin2018visual}. Raw 3D shape data are collected from the shapes in PartNet \cite{mo2019partnet} which covers common object types in typical indoor scenes. A question-answering 3D GUI is developed to collect raw point-wise annotation on mesh shapes. In total, we hired 42 professional annotators for annotations, the average annotation time per shape is 2 minutes and each shape is annotated by 3 annotators.
Finally, label propagation is employed to obtain probabilistic ground-truth for the shape point cloud.

\subsection{Affordance Type}
We refer to \cite{hassanin2018visual} for a full review of affordance types adopted in visual affordance research. 
From the full list of possible affordances, we select those suitable for 3D objects present in PartNet \cite{mo2019partnet} and remove the irrelevant ones, e.g. `reachable' and `carryable'. Overall, we filter out $18$ categories of affordances, namely `grasp', `lift', `contain', `open', `lay', `sit', `support', `wrap-grasp', `pour', `display', `push', `pull', `listen', `wear', `press', `cut', `stab', and `move'. Then, we associate the affordance types to each category of object in PartNet according to its attributes and functionality that it can afford to interact with human or robot. For example, a chair is `sittable' but not `layable', a table can afford `support' but not `contain', \etc. The affordances of each category are shown in Tab.~\ref{dataset_stats}. The annotators are allowed to freely determine where the affordance locates on the object, \eg `grasp' of bag can be annotated at its handle, webbing or straps. Notice that we allow the annotators to select the supported affordance for each shape, thus some shapes may not have all affordances defined for its own shape category. 


\subsection{Annotation Interface}

We created a web-based 3D GUI to collect raw annotations. The process of annotation is designed to be a question-answering workflow as illustrated in Fig.~\ref{figure.interface}. A user is given one shape at a time visualized in 3D. Each individual parts are colored according to the pre-defined colormap in PartNet dataset \cite{mo2019partnet}. Annotators are allowed to freely rotate, translate and change the scale of the shape
using mouse, which allow the annotators to observe the shape from more angles. After observation, annotators are first asked to determine the supported affordances by choosing from a list (`What affordances does this shape support?'). Considering that some
annotators may not understand the affordances, we provide the explanation of
each affordance on the interface. Annotators are then asked to select keypoints that support the specified affordance (`What points on the shape support the affordance?'). At least 3 keypoints will be labeled by one annotator for each affordance. Annotators will also decide whether the selected affordance will propagate beyond the part which the labeled keypoint sits on. If yes, annotators are asked to select eligible parts that the affordance can propagate to, otherwise, more annotations will be made on the same part until enough keypoints are collected.



The questions that the annotation interface proposes for each affordance directly
determine how the annotators perceive the affordances. Therefore, we define questions carefully tailored
for each affordance. Some questions are shown in Tab.~\ref{question_table}. A complete list of affordance question is given in the supplementary material.

\begin{table*}[!htb]
   \resizebox{\textwidth}{!}{%
      \begin{tabular}{c|c|c}
         \hline
         \textbf{Affordance} & \textbf{Object} & \textbf{Question}                                                                                                                                \\ \hline
         Lay             & Bed             & If you were to lie on this bed, which points would you lie on the bed?                                                                           \\
         Grasp           & Earphone        & If you want to grab this earphone, where will your palm position be?                                                                             \\
         Lift            & Bag             & If you want to lift this bag, at which points are your finger most likely to carry the bag?                                                      \\
         Sit            & Chair           & If you were sitting on this chair, on which points would you sit?                                                                                \\
         Move             & Table           & If you want to move this table, at which points on this table will you exert your strength?                                                      \\
         Open            & Trash Can       & If you want to open the lid of this trash can, from which points on the trash can you open it?                                                   \\
         Pour            & Bottle          & Suppose there is water in the bottle, and you want to pour the water out of the bottle. From which points on the bottle will the water flow out? \\
         Press           & Laptop          & If you want to press keys on a computer keyboard, which points on the keyboard would you press?                                                  \\
         Contain         & Microwave       & If you put something in the microwave, at which points in the microwave would you put the object?                                                \\
         Support             & Table           & If you want to put something on the table, at which points on the table would you put the object?                                                \\ \hline
      \end{tabular}%
   }
   \caption{Some examples of the proposed questions for affordance annotation. }
   \vspace{-0.4cm}
   \label{question_table}
\end{table*}

\begin{figure*}[!ht]
   \begin{center}
      \includegraphics[width=1.0\linewidth]{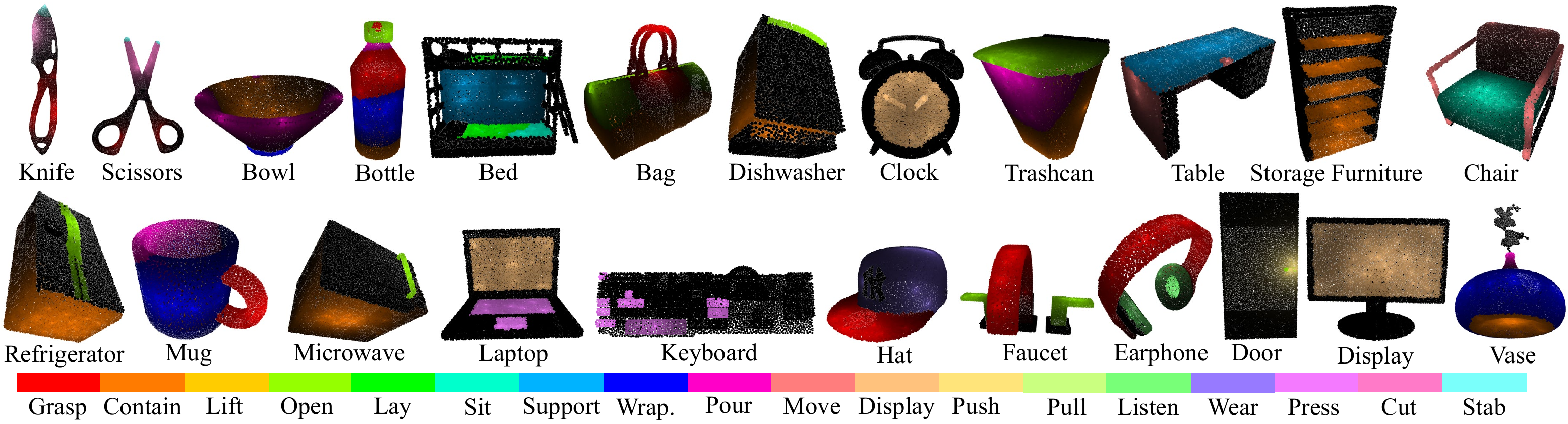}
   \end{center}
   \vspace{-0.4cm}
   \caption{The example of annotated data. Different affordances are shown in different colors, points annotated with multiple affordances are colored by the affordance that has the highest scores. The brighter the color, the higher the score.}
   \label{dataset image}
   \vspace{-0.5cm}
\end{figure*}

\subsection{Ground-Truth Construction}

After obtaining affordance keypoints, we propagate labels to all points on the shape to create ground-truth for downstream learning tasks. We first record the coordinates of the selected keypoints. We then propagate the labels to $N$ points densely sampled on the shape mesh surface, note that we only propagate on the parts that support
the specific affordance that are recorded during user annotation. Formally, we construct a kNN graph on sampled points where the adjacency matrix $\matr{A}$ writes as,
\begin{equation}
   a_{ij}=\left\{\begin{matrix}
      \left \| \vect{v}_{i}-\vect{v}_{j} \right \|_{2},\vect{v}_{j}\in NN_{k}(\vect{v}_{i}) \\
      0, \quad otherwise
   \end{matrix}\right.
   \label{math3}
\end{equation}
where $\vect{v}$ is the $xyz$ spatial coordinate of point and $NN_{k}$ denotes the set of $k$ nearest neighbors. The adjacency matrix is symmetrized by $\matr{W}=1/2(\matr{A}+\matr{A}^\top)$. Then we normalize the adjacency matrix by $\widetilde{\matr{W}}=\matr{D}^{-0.5}\matr{W}\matr{D}^{-0.5}$.
where $\matr{D}$ is the degree matrix.
Finally the scores $\matr{S}$ for all points is propagated by the closed-form solution $\matr{S}=(\matr{I}-\alpha \widetilde{\matr{W}}^{-1})\matr{Y}$.
where $\matr{Y}\in \{0,1\}^{N\times 18}$ is the one-hot label vector and 1 indicates positive label. 
$\alpha$ is a hyper-parameter controlling
the decreasing speed of $\matr{S}$, we empirically set $\alpha$ to 0.998 throughout the experiments. Finally we linearly normalize $\matr{S}$ to the range between 0 and 1 so that it is a probabilistic score. 
Example shapes with propagated affordance ground-truth are shown in Fig.~\ref{dataset image}.

\subsection{Statistics}

\begin{table}[t]
   \small
   \begin{center}
      \resizebox{\linewidth}{!}{%
         \begin{tabular}{c|c|c}
            \hline
            \textbf{Object}            & \textbf{Affordance}                                        & \textbf{Num}  \\ \hline
            \textbf{Bag}               & grasp, lift, contain, open                                 & 125                             \\
            \textbf{Bed}               & lay, sit, support                                          & 181                            \\
            \textbf{Bowl}              & contain, wrap-grasp, pour                                  & 187                        \\
            \textbf{Clock}             & display                                                    & 524                             \\
            \textbf{Dishwasher}        & open, contain                                              & 166                             \\
            \textbf{Display}           & display                                                    & 887                             \\
            \textbf{Door}              & open, push, pull                                           & 220                             \\
            \textbf{Earphone}          & grasp, listen                                              & 223                             \\
            \textbf{Faucet}            & grasp, open                                                & 628                             \\
            \textbf{Hat}               & grasp, wear                                                & 222                             \\
            \textbf{Storage Furniture} & contain, open                                              & 2186                            \\
            \textbf{Keyboard}          & press                                                      & 156                             \\
            \textbf{Knife}             & grasp, cut, stab                                           & 314                             \\
            \textbf{Laptop}            & display, press                                             & 421                            \\
            \textbf{Microwave}         & open, contain, support                                     & 184                             \\
            \textbf{Mug}               & contain, pour, wrap-grasp, grasp                           & 190                            \\
            \textbf{Refrigerator}      & contain, open                                              & 185                             \\
            \textbf{Chair}             & sit, support, move                                         & 6113                          \\
            \textbf{Scissors}          & grasp, cut, stab                                           & 68                              \\
            \textbf{Table}             & support, move                                              & 7990                            \\
            \textbf{Trash Can}         & contain, pour, open                                        & 315                            \\
            \textbf{Vase}              & contain, pour, wrap-grasp                                  & 1048                           \\
            \textbf{Bottle}            & contain, open, wrap-grasp, grasp, pour                     & 411                \\ \hline
         \end{tabular}%
      }
   \end{center}
   \vspace{-0.2cm}
   \caption{3D AffordanceNet statistics. The first column shows the object category. The second column shows the defined affordance classes for each category.The third column shows the amount of each shape semantic category. }
   \label{dataset_stats}
   \vspace{-0.2cm}
\end{table}


The final dataset provides well-defined visual affordance score map annotations for 22949 shapes from 23 shape categories with at most 5 affordance types defined for each category. From the perspective of affordance categories, 3D AfffodanceNet contains 56307 affordance annotations from 18 affordance classes. 
It is worth noting that due to the multi-label nature, each point could be labeled with multiple affordances. 
More details of the dataset are presented in Tab.~\ref{dataset_stats} and Tab.~\ref{affordance_stats}.

\section{Tasks and Benchmarks}
In this section, we benchmark three tasks to demonstrate the 3D AfffodanceNet dataset, namely, full-shape, partial-view and rotation-invariant affordance estimation. The 3D AffordanceNet dataset is split into train, validation and test sets with a ratio of 70\%, 10\% and 20\%, respectively according to the shape semantic category. 
The first experiment estimates point-wise affordance given full 3D point cloud as input. The second experiment estimates the affordance of partially visible objects observed from different viewpoints. The last experiment estimates the affordance of rotated 3D objects under two different rotation settings. We also create a semi-supervised affordance estimation benchmark to explore the opportunity of exploiting unlabeled data for affordance estimation.
\vspace{-0.1cm}

\begin{table}[t]
\resizebox{\linewidth}{!}{%
\begin{tabular}{c|ccccccccc}
\hline
\textbf{}    & \textbf{Support} & \textbf{Move} & \textbf{Sit}  & \textbf{Contain} & \textbf{Open}   & \textbf{Grasp} & \textbf{Pour} & \textbf{Display} & \textbf{Wrap-Grasp} \\ \hline
\textbf{\#Annot} & 14848            & 14540         & 6516          & 5155             & 4506            & 2253           & 2086          & 1914             & 1889                \\ \hline
\textbf{}    & \textbf{Press}   & \textbf{Cut}  & \textbf{Stab} & \textbf{Wear}    & \textbf{Listen} & \textbf{Pull}  & \textbf{Push} & \textbf{Lay}     & \textbf{Lift}       \\ \hline
\textbf{\#Annot} & 588              & 393           & 393           & 231              & 228             & 225            & 225           & 194              & 123                 \\ \hline
\end{tabular}%
}
\caption{The number of shapes that are positive for each category of affordance.}
\vspace{-0.6cm}
\label{affordance_stats}
\end{table}

\begin{table*}
   \begin{center}
      \resizebox{\textwidth}{!}{%
         \begin{tabular}{c|c|cccccccccccccccccc}
\hline
\textbf{Full-shape}   & \textbf{Avg}  & \textbf{Grasp} & \textbf{Lift} & \textbf{Contain} & \textbf{Open} & \textbf{Lay}  & \textbf{Sit}  & \textbf{Support} & \textbf{Wrap.} & \textbf{Pour} & \textbf{Display} & \textbf{Push} & \textbf{Pull} & \textbf{Listen} & \textbf{Wear} & \textbf{Press} & \textbf{Move} & \textbf{Cut}  & \textbf{Stab} \\ \hline
\textbf{P\_mAP}       & 48.0          & 43.4           & 75.1          & 56.9             & 46.6          & 63.0          & 81.1          & 52.5             & 19.0           & 46.9          & 59.3             & 20.5          & 37.9          & 41.6            & 20.4          & 31.4           & 35.3          & 41.1          & 90.9          \\
\textbf{P\_AUC}       & 87.4          & 82.8           & 97.1          & 89.3             & 90.6          & 92.6          & 96.0          & 89.7             & 72.6           & 89.2          & 90.6             & 83.1          & 85.3          & 85.9            & 67.9          & 90.9           & 79.0          & 91.4          & 98.8          \\
\textbf{P\_aIOU}      & 19.3          & 15.7           & 41.2          & 22.2             & 20.2          & 30.0          & 38.1          & 17.5             & 4.0            & 18.2          & 25.6             & 6.5           & 12.7          & 14.0            & 6.5           & 11.2           & 8.5           & 15.2          & 40.9          \\
\textbf{P\_MSE}       & 0.059         & 0.003          & 0.0001        & 0.006            & 0.003         & 0.0005        & 0.005         & 0.012            & 0.002          & 0.002         & 0.002            & 0.0007        & 0.0002        & 0.0006          & 0.0005        & 0.0006         & 0.021         & 0.0003        & 0.0001        \\
\textbf{D\_mAP}       & 46.4          & 43.9           & 85.2          & 57.6             & 51.8          & 12.3          & 80.9          & 54.0             & 20.7           & 47.7          & 65.5             & 20.5          & 40.5          & 36.0            & 18.3          & 34.2           & 35.5          & 40.2          & 91.4          \\
\textbf{D\_AUC}       & 85.5          & 82.5           & 98.7          & 89.9             & 91.6          & 50.1          & 96.1          & 90.2             & 74.6           & 89.2          & 92.1             & 85.0          & 89.7          & 86.1            & 61.9          & 91.8           & 78.9          & 91.7          & 98.7          \\
\textbf{D\_aIOU}      & 17.8          & 13.9           & 40.2          & 21.6             & 25.4          & 1.0           & 34.9          & 18.8             & 5.6            & 17.7          & 32.1             & 5.5           & 11.8          & 11.9            & 5.9           & 14.8           & 9.9           & 14.5          & 35.4          \\
\textbf{D\_MSE}       & 0.08          & 0.003          & 0.0001        & 0.007            & 0.003         & 0.0006        & 0.006         & 0.013            & 0.007          & 0.005         & 0.002            & 0.002         & 0.0006        & 0.002           & 0.002         & 0.0007         & 0.025         & 0.0002        & 0.0001        \\
\textbf{U\_mAP}       & 47.4          & 42.6           & 75.7          & 56.6             & 45.9          & 60.6          & 80.8          & 53.7             & 19.1           & 45.0          & 61.5             & 19.7          & 36.5          & 37.4            & 20.4          & 33.9           & 35.2          & 39.8          & 89.0          \\
\textbf{U\_AUC}       & 86.3          & 79.8           & 94.3          & 88.8             & 88.5          & 88.2          & 95.8          & 89.7             & 72.9           & 87.1          & 90.2             & 81.4          & 84.0          & 82.9            & 70.3          & 91.6           & 79.2          & 90.8          & 98.5          \\
\textbf{U\_aIOU}      & 19.7          & 13.7           & 41.2          & 22.4             & 20.8          & 29.4          & 37.3          & 18.6             & 4.7            & 18.3          & 32.4             & 6.2           & 13.0          & 13.1            & 4.4           & 14.5           & 9.2           & 14.2          & 41.5          \\
\textbf{U\_MSE}       & 0.063         & 0.003          & 0.0003        & 0.006            & 0.003         & 0.0006        & 0.005         & 0.014            & 0.002          & 0.002         & 0.002            & 0.0006        & 0.0002        & 0.0007          & 0.0003        & 0.0008         & 0.021         & 0.0002        & 0.0001        \\ \hline
\textbf{Partial}      & \textbf{Avg}  & \textbf{Grasp} & \textbf{Lift} & \textbf{Contain} & \textbf{Open} & \textbf{Lay}  & \textbf{Sit}  & \textbf{Support} & \textbf{Wrap.} & \textbf{Pour} & \textbf{Display} & \textbf{Push} & \textbf{Pull} & \textbf{Listen} & \textbf{Wear} & \textbf{Press} & \textbf{Move} & \textbf{Cut}  & \textbf{Stab} \\ \hline
\textbf{P\_mAP}       & 45.7          & 43.2           & 80.6          & 41.9             & 48.5          & 52.6          & 69.8          & 45.5             & 20.0           & 47.0          & 52.5             & 24.1          & 36.5          & 42.1            & 15.3          & 30.7           & 37.8          & 43.3          & 92.6          \\
\textbf{P\_AUC}       & 85.2 & 81.2  & 96.2 & 83.3    & 87.9 & 86.7 & 95.0 & 86.5    & 71.3  & 88.4 & 85.2    & 84.9 & 86.1 & 84.2   & 64.1 & 84.7  & 79.6 & 90.2 & 98.8 \\
\textbf{P\_aIOU}      & 16.9          & 14.4           & 45.6          & 13.2             & 21.6          & 25.2          & 31.0          & 11.2             & 3.6            & 17.8          & 19.3             & 5.7           & 11.5          & 13.4            & 2.4           & 12.4           & 6.2           & 13.5          & 37.8          \\
\textbf{P\_MSE}       & 0.062         & 0.003          & 0.0001        & 0.005            & 0.003         & 0.0006        & 0.004         & 0.013            & 0.002          & 0.002         & 0.002            & 0.0002        & 0.0001        & 0.0007          & 0.0004        & 0.0006         & 0.025         & 0.0003        & 0.0001        \\
\textbf{D\_mAP}       & 42.2          & 40.1           & 83.8          & 38.5             & 44.5          & 46.6          & 67.3          & 43.9             & 19.6           & 44.8          & 50.3             & 15.7          & 28.6          & 19.9            & 17.7          & 26.9           & 35.6          & 43.3          & 92.1          \\
\textbf{D\_AUC}       & 83.7          & 80.3           & 97.0          & 80.8             & 86.8          & 82.6          & 94.6          & 87.0             & 70             & 86.8          & 81.7             & 83.6          & 86.0          & 75              & 64.1          & 83.0           & 78.5          & 90.0          & 99.1          \\
\textbf{D\_aIOU}      & 13.8          & 13.3           & 37.9          & 8.6              & 16.0          & 15.5          & 27.2          & 13.5             & 3.2            & 13.9          & 13.8             & 3.9           & 4.3           & 8.4             & 5.2           & 5.7            & 8.2           & 13.3          & 36.8          \\
\textbf{D\_MSE}       & 0.069         & 0.005          & 0.0002        & 0.005            & 0.004         & 0.0006        & 0.004         & 0.013            & 0.003          & 0.002         & 0.002            & 0.001         & 0.001         & 0.002           & 0.003         & 0.0004         & 0.022         & 0.0004        & 0.0001        \\
\textbf{U\_mAP}       & 43.0          & 40.8           & 73.2          & 41.2             & 47.6          & 44.6          & 68.9          & 45.3             & 18.5           & 45.2          & 53.2             & 19.5          & 29.6          & 38.9            & 11.1          & 29.6           & 36.7          & 39.1          & 90.8          \\
\textbf{U\_AUC}       & 83.2          & 78.9           & 94.6          & 81.7             & 87.1          & 80.4          & 94.5          & 87.0             & 69.3           & 87.5          & 84.4             & 80.9          & 79.2          & 84.2            & 56.4          & 84.9           & 77.7          & 89.3          & 98.7          \\
\textbf{U\_aIOU}      & 16.8          & 14.7           & 38.1          & 15.0             & 21.1          & 20.8          & 33.5          & 14.4             & 4.3            & 18.9          & 20.4             & 5.3           & 8.2           & 10.9            & 1.0           & 14.7           & 9.0           & 15.9          & 35.8          \\
\textbf{U\_MSE}       & 0.065         & 0.003          & 0.0002        & 0.006            & 0.003         & 0.001         & 0.006         & 0.012            & 0.003          & 0.003         & 0.002            & 0.0003        & 0.0002        & 0.0003          & 0.0006        & 0.0008         & 0.022         & 0.0007        & 0.0002        \\ \hline
\textbf{Rotate z}     & \textbf{Avg}  & \textbf{Grasp} & \textbf{Lift} & \textbf{Contain} & \textbf{Open} & \textbf{Lay}  & \textbf{Sit}  & \textbf{Support} & \textbf{Wrap.} & \textbf{Pour} & \textbf{Display} & \textbf{Push} & \textbf{Pull} & \textbf{Listen} & \textbf{Wear} & \textbf{Press} & \textbf{Move} & \textbf{Cut}  & \textbf{Stab} \\ \hline
\textbf{P\_mAP}       & 47.3          & 43.1           & 85.7          & 58.1             & 39.6          & 62.7          & 80.6          & 53.8             & 20.4           & 47.5          & 47.2             & 21.8          & 34.8          & 39.7            & 19.0          & 28.1           & 36.3          & 40.4          & 91.9          \\
\textbf{P\_AUC}       & 87.0          & 82.0           & 97.9          & 89.5             & 86.2          & 91.1          & 95.9          & 90.1             & 74.2           & 89.4          & 87.1             & 85.4          & 87.9          & 84.3            & 67.0          & 88.5           & 80.3          & 91.5          & 98.5          \\
\textbf{P\_aIOU}      & 18.7          & 15.5           & 45.4          & 22.4             & 17.6          & 26.0          & 38.0          & 18.3             & 4.6            & 19.5          & 17.4             & 7.0           & 10.5          & 13.9            & 6.9           & 9.2            & 8.8           & 14.8          & 40.6          \\
\textbf{P\_MSE}       & 0.06          & 0.003          & 0.0001        & 0.006            & 0.003         & 0.0006        & 0.005         & 0.012            & 0.002          & 0.002         & 0.003            & 0.001         & 0.0003        & 0.0007          & 0.0007        & 0.0005         & 0.02          & 0.0003        & 0.0001        \\
\textbf{D\_mAP}       & 44.8          & 42.2           & 82.9          & 58.2             & 45.2          & 17.3          & 78.8          & 52.4             & 20.3           & 46.7          & 58.5             & 21.6          & 45.2          & 28.5            & 16.8          & 29.6           & 35.0          & 36.4          & 90.8          \\
\textbf{D\_AUC}       & 84.9          & 80.8           & 98.2          & 89.9             & 87.9          & 54.9          & 95.8          & 89.9             & 74.0           & 89.5          & 90.6             & 84.7          & 89.4          & 81.0            & 64.4          & 89.2           & 79.7          & 90.6          & 98.4          \\
\textbf{D\_aIOU}      & 16.1          & 13.6           & 36.1          & 19.6             & 21.2          & 1.0           & 29.2          & 18.5             & 3.4            & 13.6          & 25.3             & 5.7           & 13.6          & 11.3            & 5.7           & 13.1           & 9.7           & 13.6          & 36.3          \\
\textbf{D\_MSE}       & 0.074         & 0.003          & 0.0001        & 0.007            & 0.003         & 0.0006        & 0.007         & 0.013            & 0.005          & 0.004         & 0.002            & 0.003         & 0.0008        & 0.002           & 0.002         & 0.0009         & 0.021         & 0.0004        & 0.0001        \\
\textbf{U\_mAP}       & 46.1          & 42.7           & 74.4          & 56.1             & 40.4          & 58.4          & 81.2          & 54.9             & 18.5           & 44.7          & 56.4             & 20.7          & 35.3          & 36.8            & 17.4          & 31.6           & 35.6          & 36.2          & 88.8          \\
\textbf{U\_AUC}       & 86.1          & 81.2           & 95.8          & 87.5             & 85.9          & 88.4          & 95.8          & 90.2             & 72.2           & 87.6          & 87.9             & 85.1          & 87.9          & 83.2            & 63.1          & 90.0           & 78.9          & 90.3          & 98.4          \\
\textbf{U\_aIOU}      & 18.9          & 15.5           & 39.6          & 21.8             & 18.4          & 24.7          & 38.4          & 18.9             & 4.5            & 18.6          & 26.9             & 6.6           & 12.2          & 14.5            & 5.1           & 14.1           & 9.9           & 12.1          & 37.7          \\
\textbf{U\_MSE}       & 0.06          & 0.003          & 0.0001        & 0.006            & 0.003         & 0.0004        & 0.005         & 0.013            & 0.002          & 0.002         & 0.002            & 0.0007        & 0.0002        & 0.0008          & 0.0004        & 0.0008         & 0.021         & 0.0003        & 0.0001        \\ \hline
\textbf{Rotate SO(3)} & \textbf{Avg}  & \textbf{Grasp} & \textbf{Lift} & \textbf{Contain} & \textbf{Open} & \textbf{Lay}  & \textbf{Sit}  & \textbf{Support} & \textbf{Wrap.} & \textbf{Pour} & \textbf{Display} & \textbf{Push} & \textbf{Pull} & \textbf{Listen} & \textbf{Wear} & \textbf{Press} & \textbf{Move} & \textbf{Cut}  & \textbf{Stab} \\ \hline
\textbf{P\_mAP}       & 41.8          & 40.5           & 78.5          & 42.4             & 32.7          & 38.4          & 74.5          & 48.3             & 19.4           & 41.7          & 41.1             & 19.7          & 30.3          & 39.4            & 17.6          & 21.5           & 34.6          & 41.1          & 90.6          \\
\textbf{P\_AUC}       & 83.3          & 79.0           & 93.6          & 81.1             & 81.3          & 79.6          & 93.9          & 87.4             & 71.6           & 85.4          & 83.7             & 83.1          & 84.0          & 84.2            & 64.8          & 78.2           & 78.6          & 91.0          & 99.4          \\
\textbf{P\_aIOU}      & 15.2          & 12.8           & 38.3          & 12.2             & 13.1          & 9.6           & 33.6          & 16.5             & 3.8            & 16.1          & 13.7             & 3.0           & 11.1          & 14.8            & 5.5           & 8.8            & 8.9           & 14.4          & 37.2          \\
\textbf{P\_MSE}       & 0.072         & 0.003          & 0.0001        & 0.008            & 0.003         & 0.0007        & 0.007         & 0.015            & 0.003          & 0.003         & 0.003            & 0.0002        & 0.0001        & 0.0006          & 0.002         & 0.0009         & 0.022         & 0.0006        & 0.0001        \\
\textbf{D\_mAP}       & 37.3          & 37.9           & 70.7          & 37.3             & 34.2          & 9.7           & 73.9          & 46.8             & 17.6           & 40.2          & 46.8             & 19.1          & 37.4          & 6.9             & 11.8          & 22.3           & 30.9          & 41.4          & 86.4          \\
\textbf{D\_AUC}       & 78.9          & 79.1           & 95.6          & 79.4             & 81.4          & 36.3          & 93.9          & 87.2             & 71.2           & 85.6          & 85.7             & 82.3          & 88.8          & 43.5            & 60.3          & 82.7           & 76.9          & 91.9          & 99.1          \\
\textbf{D\_aIOU}      & 12.8          & 13.6           & 32.5          & 7.8              & 13.9          & 1.0           & 35.4          & 14.4             & 4.9            & 16.4          & 19.3             & 4.5           & 11.0          & 0.003           & 1.0           & 8.4            & 7.1           & 15.7          & 23.3          \\
\textbf{D\_MSE}       & 0.08          & 0.004          & 0.0002        & 0.007            & 0.003         & 0.0006        & 0.007         & 0.015            & 0.005          & 0.004         & 0.005            & 0.002         & 0.0005        & 0.0004          & 0.0006        & 0.0009         & 0.022         & 0.0007        & 0.0002        \\
\textbf{U\_mAP}       & 37.9          & 37.0           & 61.2          & 38.0             & 29.8          & 34.0          & 77.4          & 49.9             & 16.4           & 39.3          & 42.6             & 14.8          & 24.7          & 35.7            & 8.6           & 20.1           & 31.8          & 36.9          & 83.3          \\
\textbf{U\_AUC}       & 80.9          & 76.9           & 90.5          & 79.5             & 77.7          & 78.1          & 94.1          & 87.8             & 67.9           & 82.7          & 81.9             & 78.1          & 83.5          & 83.7            & 52.5          & 76.6           & 77.1          & 89.6          & 99.0          \\
\textbf{U\_aIOU}      & 12.0          & 15.3           & 8.2           & 10.8             & 10.7          & 5.4           & 35.8          & 16.2             & 2.7            & 12.7          & 15.8             & 1.0           & 4.3           & 12.3            & 1.0           & 7.3            & 7.9           & 11.9          & 35.8          \\
\textbf{U\_MSE}       & 0.07          & 0.005          & 0.0001        & 0.008            & 0.003         & 0.0005        & 0.006         & 0.013            & 0.003          & 0.003         & 0.004            & 0.0002        & 0.0001        & 0.0004          & 0.0004        & 0.0008         & 0.022         & 0.0003        & 0.0001        \\ \hline
\end{tabular}%
      }
   \end{center}
   \vspace{-0.1cm}
   \caption{Affordance Estimation Results. Except for the MSE results, others are shown in percentage, the higher the scores the higher the results. Algorithm P, D and U represent PointNet++\cite{qi2017pointnet++}, DGCNN\cite{wang2019dynamic} and U-Net\cite{xie2020pointcontrast} respectively. The words \textit{Full-Shape}, \textit{Partial}, \textit{Rotate z} and \textit{Rotate SO(3)} represent the full-shape, partial, $z/z$ and $SO(3)/SO(3)$ rotation-invariant affordance estimation, respectively.}
   \label{Estimation results}
   \vspace{-0.5cm}
\end{table*}

\begin{figure*}[t]
   \begin{center}
      \includegraphics[width=1.0\textwidth]{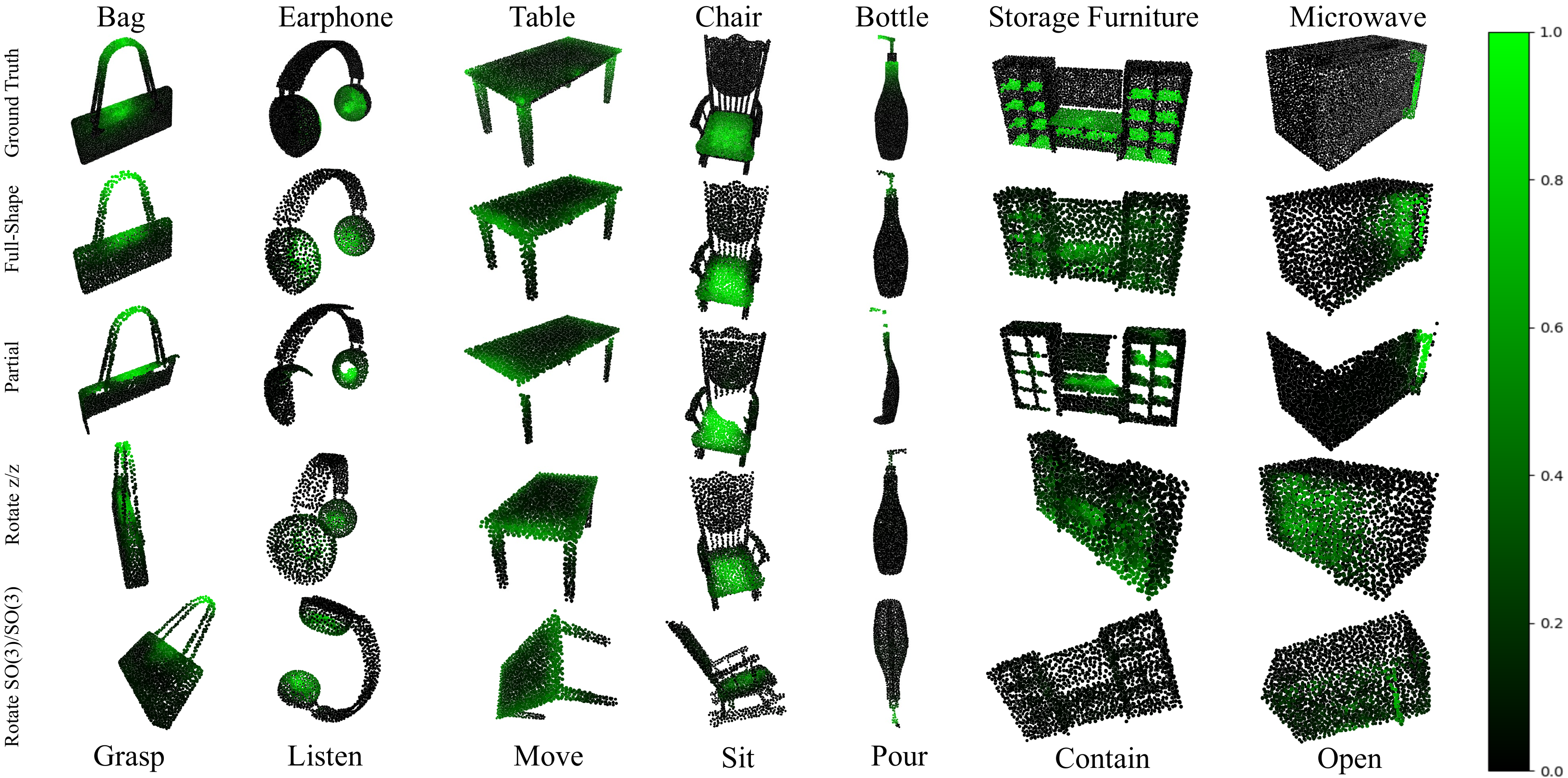}
   \end{center}
   \vspace{-0.3cm}
   \caption{Qualitative results for affordance estimation. The top row shows the ground truth. The second row shows the full-shape estimated results, the third row shows the partial-view estimated result, the fourth and the bottom row show the $z/z$ and $SO(3)/SO(3)$ rotation-invariant estimated results, respectively. All results come from PointNet++. The top words indicate the semantic category of each column and the bottom words indicate the affordance category. The greener the color of the points, the higher the confidence about specific affordance types. \textit{Wrap.} is the abbreviation of \textit{Wrap-Grasp.}}
   \label{estimation image}
   \vspace{-0.5cm}
\end{figure*}

\subsection{Full-Shape Affordance Estimation} \label{estimation task}

Given an object as 3D point cloud without knowing the affordances supported by the object, the full-shape affordance estimation task aims to estimate the supported affordance type and predict the point-wise probabilistic score of affordance. We show that state-of-the-art 3D point cloud segmentation networks predict reasonable results on 3D AffordanceNet. 

\noindent\textbf{Network and Training.} {We evaluate three network architecture, namely PointNet++\cite{qi2017pointnet++}, DGCNN\cite{wang2019dynamic} and U-Net\cite{xie2020pointcontrast} for this task.} To obtain the point-wise score, we utilized the segmentation branch of PointNet++ and DGCNN as shared backbones to extract features for each point, then for each affordance type, we pass the features through multiple classification heads and used a sigmoid function to obtain the posterior scores. The classification heads were set up for each affordance category individually while the backbone networks were shared. We use cross-entropy loss $L_{CE}$ for training the network as below,
\begin{equation}
    \resizebox{\linewidth}{!}{
    $l_{CE} = \frac{1}{N}\sum_{i}^{M}\sum_{j}^{N}-(1-t_{ij})log(1-p_{ij})-t_{ij}log(p_{ij})$
    }
\end{equation}
where $M$ is the total number of affordance types, $N$ is  the number of points within each shape, $s_{ij}$ is the ground truth score of $j$th point of $i$th affordance category and $p_{ij}$ is the predicted score.

Since the points with zero score account for a relatively large proportion of the total dataset, we further propose to use dice loss \cite{milletari2016v} to mitigate the imbalance issue. The dice loss $l_{DICE}$ is defined as:
\vspace{-0.2cm}
\begin{equation}
    \begin{split}
    l_{DICE} = \sum_{i}^{M}1 - \frac{\sum_{j}^{N}s_{i,j}p_{i,j}+\epsilon}{\sum_{j}^{N}s_{i,j}+p_{i,j}+\epsilon} \\
      -\frac{\sum_{j}^{N}(1-s_{i,j})(1-p_{i,j})+\epsilon}{\sum_{j}^{N}2 - s_{i,j} - p_{i,j}+\epsilon}
    \end{split}
\end{equation}
Finally the loss function is defined as $l = l_{CE}+l_{DICE}$. {We train PointNet++ and DGCNN on the 3D AffordanceNet using the default training strategies and hyper-parameters described in respective papers\cite{qi2017pointnet++,wang2019dynamic}. For Unet, we fine-tune the network initialized by the pre-trained weight provided by PointContrast\cite{xie2020pointcontrast}.}

\noindent\textbf{Evaluation and Results.} We evaluate four metrics for affordance estimation, including mean Average Precision (mAP) scores, {mean squared error (MSE), }Area Under ROC Curve (AUC) and average Intersection Over Union (aIoU). {For AP, we calculate the Precision-Recall Curve and AP is calculated for each affordance. For AUC, we report the area under ROC Curve. For MSE, we calculate mean squared error of each affordance category and sum up the results from all affordance categories. }For aIoU, we gradually tune up the threshold from 0 to 0.99 with 0.01 step to binarize the prediction, and the aIoU is the arithmetic average of all IoUs at each threshold. {Except for the MSE, all the other metrics }for each category are averaged over all shapes, a.k.a. macro-average. For each affordance category, the ground-truth map is binarized with 0.5 threshold before evaluation. The results are reported in Tab.~\ref{Estimation results} under the Full-Shape section and some qualitative examples from PointNet++ are selected and visualized in Fig.~\ref{estimation image}.


As shown in Tab.~\ref{Estimation results}, the performances of three networks are close and all achieve a relatively low aIOU score, which indicates that affordance estimation is still a challenging task. Comparing the second row of Fig.~\ref{estimation image} to corresponding ground truth, we found that PointNet++ produces some reasonable results. For example, the estimations of \textit{grasp} on a \textit{bag} are successfully localized on both the handles and the webbing. However, the results of \textit{pour} on a \textit{bottle} fail since the network predicts the scores mainly on the lid of the bottle rather than the body edge of the bottle where is the place that the water flow out. More qualitative examples are given in the supplementary.

\noindent\textbf{Room for Performance Improvement} \label{improvement}
The performances of PointNet++ and DGCNN on the tasks mentioned above are relatively weak. Hence, we evaluate the trained network on full-shape affordance estimation task over the training, validation and testing sets to investigate the room for performance improvement with results reported in Tab.~\ref{Set_Performances}. From the results we observe that both two networks still under-fit, meaning that the proposed affordance estimation task is very challenging for existing point cloud analysis networks.

\begin{table}[!htb]
\resizebox{\linewidth}{!}{%
\begin{tabular}{c|ccccc|c|ccccc}
\hline
        & \textbf{mAP}  & \textbf{AUC}  & \textbf{aIOU} & \textbf{MSE}   & \textbf{Loss}     &         & \textbf{mAP}  & \textbf{AUC}  & \textbf{aIOU} & \textbf{MSE}   & \textbf{Loss}     \\ \hline
\textbf{P Train} & 52.3 & 89.8 & 21.7 & 0.054 & 8.75 & \textbf{D Train} & 51.7 & 89.1 & 21.2 & 0.061 & 8.83 \\
\textbf{P Val}   & 48.2 & 88.0   & 19.2 & 0.057 & 8.81 & \textbf{D Val}   & 47.8 & 85.8 & 17.5 & 0.075 & 8.91 \\
\textbf{P Test}  & 48.0   & 87.4 & 19.3 & 0.059 & 8.83 & \textbf{D Test}  & 46.4 & 85.5 & 17.8 & 0.08  & 8.93 \\ \hline
\end{tabular}%
}
\caption{The performances of two different networks on full-shape affordance estimation task over train, validate and test sets. P represents PointNet++ and D refers to DGCNN.}
\label{Set_Performances}
\end{table}

\subsection{Partial-View Affordance Estimation} \label{partial task}

Although complete point clouds or meshes can provide detailed geometric information for affordance estimation, in real-world application scenarios, we can only expect partial view of 3D shapes, represented as partial point cloud. Therefore, another important task we are concerned with is to estimate the affordance from partial point cloud.

\noindent\textbf{Network and Training.} To obtain partial point clouds, we follow \cite{katz2007direct} to synthesize point cloud observed from certain camera viewpoints. Only points directly facing the camera will be preserved as visible points and each point is assigned a radius to create occlusion effect. In specific, because all shapes are well aligned within the (-1,-1,-1) to (1,1,1) cube, we set up 4 affine cameras located at (1,1,1), (-1,-1,1), (1,-1,-1), (-1,1,-1) in Cartesian coordinate system, facing towards the origin. After obtaining the partial point clouds, we sample 2048 points from each viewpoint via furthest point sampling, if the number of points of the point cloud is fewer than 2048, we utilize the point cloud up-sampling method proposed in \cite{Yifan_2019_CVPR} to up-sample the data. We use exactly the same backbone networks and training strategies described in previous sections.

\noindent\textbf{Evaluation and Results.} During testing stage, we estimate the affordance on the visible partial point cloud only. The evaluation protocol follows the one described in Section~\ref{estimation task}. All evaluation metrics are reported in Tab.~\ref{Estimation results} with qualitative results from PointNet++ shown in Fig.~\ref{estimation image}. 


Unsurprisingly, the quantitative performances of three networks have decreased due to the loss of geometric information of partial point cloud relative to complete point cloud. Nevertheless, we still observe reasonable qualitative results even though only a partial view is observed. For instance, the network produces high prediction for \textit{move} on the upside of the legs of a \textit{table} despite the unseen parts of the legs. The \textit{grasp} for \textit{bag}, \textit{hear} for \textit{earphone}, \textit{sit} for \textit{chair}, etc., are all more-or-less correctly predicted. In contrast, the estimation for \textit{contain} on \textit{storage furniture} are partially missing since it predicts the scores on the top of the furniture which is not fully observed. 

\begin{table*}[t]
   \begin{center}
      \resizebox{\textwidth}{!}{%
         \begin{tabular}{c|c|cccccccccccccccccc}
\hline
\textbf{VAT} & \textbf{Avg} & \textbf{Grasp} & \textbf{Lift} & \textbf{Contain} & \textbf{Open} & \textbf{Lay} & \textbf{Sit} & \textbf{Support} & \textbf{Wrap.} & \textbf{Pour} & \textbf{Display} & \textbf{Push} & \textbf{Pull} & \textbf{Listen} & \textbf{Wear} & \textbf{Press} & \textbf{Move} & \textbf{Cut} & \textbf{Stab} \\ \hline
\textbf{mAP}        & 36.6         & 34.8           & 78.8          & 44.3             & 22.6          & 34.9         & 64.8         & 41.4             & 13.3           & 41.5          & 58.9             & 13.5          & 8.3           & 20.5            & 8.5           & 29.5           & 19.4          & 33.1         & 90.2          \\
\textbf{AUC}        & 78.8         & 75.8           & 95.3          & 78.4             & 79.2          & 71.2         & 92.4         & 86.7             & 56.5           & 83.7          & 89.9             & 69.0            & 74.8          & 74.1            & 51.4          & 88.8           & 62.7          & 89.3         & 99.1          \\
\textbf{aIOU}       & 11.2         & 12.7           & 20.3          & 14.2             & 4.8           & 6.4          & 16.9         & 7.7              & 5.1            & 17.6          & 30.2             & 2.5           & 1.1           & 8.4             & 3.6           & 12.8           & 2.9           & 9.5          & 24.3          \\
\textbf{MSE}        & 0.155        & 0.007          & 0.0001        & 0.02             & 0.006         & 0.0007       & 0.012        & 0.023            & 0.023          & 0.007         & 0.008            & 0.0003        & 0.0001        & 0.002           & 0.002         & 0.005          & 0.034         & 0.004        & 0.0003        \\ \hline
\textbf{Full-Shape}        & \textbf{Avg} & \textbf{Grasp} & \textbf{Lift} & \textbf{Contain} & \textbf{Open} & \textbf{Lay} & \textbf{Sit} & \textbf{Support} & \textbf{Wrap.} & \textbf{Pour} & \textbf{Display} & \textbf{Push} & \textbf{Pull} & \textbf{Listen} & \textbf{Wear} & \textbf{Press} & \textbf{Move} & \textbf{Cut} & \textbf{Stab} \\ \hline
\textbf{mAP}        & 34.3         & 34.5           & 77.9          & 42.3             & 18.2          & 36.5         & 62.9         & 38.8             & 11.2           & 38.9          & 53.3             & 13.4          & 7.3           & 9               & 6.7           & 24.5           & 17.4          & 35.8         & 89.5          \\
\textbf{AUC}        & 77.5         & 75.4           & 94.8          & 78.1             & 75.7          & 73.3         & 92.1         & 85.8             & 54.4           & 82.1          & 87.5             & 74.5          & 77.4          & 61.6            & 46.7          & 83.4           & 64.1          & 89.8         & 98.9          \\
\textbf{aIOU}       & 9.8          & 11.2           & 28.1          & 14.3             & 2.5           & 10.0           & 23.4         & 9.8              & 2.2            & 7.5           & 19.9             & 1.9           & 1.0             & 1.6             & 1.6           & 6.8            & 2.3           & 5.6          & 26.5          \\
\textbf{MSE}        & 0.105        & 0.009          & 0.0002        & 0.013            & 0.003         & 0.001        & 0.013        & 0.021            & 0.004          & 0.003         & 0.003            & 0.0002        & 0.0001        & 0.0004          & 0.001         & 0.0008         & 0.031         & 0.0007       & 0.0001        \\ \hline
\end{tabular}%
      }
   \end{center}
   \vspace{-0.2cm}
   \caption{The Results of Semi-Supervised Affordance Estimation. All numbers are in $\%$ except for MSE. We only implement semi-supervised affordance estimation on DGCNN. The words \textit{Full-Shape} and \textit{VAT} represent full-shape estimation and semi-supervised affordance estimation with virtual adversarial training. \textit{Wrap.} is the abbreviation of \textit{Wrap-Grasp.}}
   \label{Semi results}
   \vspace{-0.4cm}
\end{table*}

\subsection{Rotation-Invariant Affordance Estimation} \label{rotate task}

The shapes in 3D AffordanceNet are all aligned in canonical poses, however, the data observed by sensors in real world are not always in canonical poses. The difference in rotation between real data and training data will lead to a performance drop in real-world usage which inspired research into rotation equivariant network \cite{esteves2018learning}. Hence, it is critical to train the algorithms to estimate affordance on rotated objects. In this section, we provide a benchmark for affordance estimation subject to two types of rotations.

\noindent\textbf{Network and Training.} We used the same backbone networks, training strategies and hyper-parameters described in the Sect.~\ref{estimation task}. We propose two different rotation settings for experiment: $z/z$ and $SO(3)/SO(3)$ where $z/z$ means rotation is applied along z axis only for both training and inference stages while $SO(3)/SO(3)$ refers to $SO(3)$ rotation, i.e. freely rotation along x, y and z axes. During training session, we randomly sample rotation poses between [0,2$\pi$] for each shape in the training mini-batch on-the-fly. We train proposed methods on complete point cloud. For testing phase, we randomly sample 5 rotation poses for each shape for both rotation settings and fix the sampled rotations for testing data.

\noindent\textbf{Evaluation and Results.} We calculate mAP, AUC, aIOU on the proposed methods. Quantitative results are presented in Tab.~\ref{Estimation results} and qualitative results are shown in the fourth and fifth rows of Fig.~\ref{estimation image} for $z/z$ and $SO(3)/SO(3)$ settings with PointNet++ as backbone. We observe that the performances on both $z/z$ and $SO(3)/SO(3)$ settings dropped compared to canonical view experiments. In particular, for $z/z$ setting, the performance dropped around $1\%$ in all metrics for all networks as backbone. While more significant loss of performance is observed for $SO(3)/SO(3)$ setting with $5-10\%$ drop in all metrics. This is aligned with our expectation that $SO(3)/SO(3)$ is a much more challenging task. We further make observations from the qualitative results in Fig.~\ref{estimation image}. First, under $z/z$ rotation scheme, despite the consistent performance drop, affordance estimations are largely correct across most categories. Obvious mistakes are made in \textit{bottle} and \textit{microwave} where the former missed the tip which supports \textit{pour} while the latter mistakenly predict the while door of \textit{microwave} for \textit{open}. Under the more challenging $SO(3)/SO(3)$ scheme, there is still a visually satisfying results for most shapes. The most prominent error is made on \textit{storage furniture} where \textit{contain} is totally missed, probably because the complex geometric structure (many concave shapes) renders \textit{contain} a hard affordance to learn under arbitrary rotation.
In general, we believe affordance estimation under $SO(3)/SO(3)$ a very challenging task and it deserves further investigation.


\subsection{Semi-Supervised Affordance Estimation}

Although the label propagation procedure allows the annotators to only annotate a few keypoints on the object surface, affordance annotation still remains as an expensive and labor intensive procedure. Inspired by the recent success in semi-supervised learning (SSL) \cite{laine2016temporal,tarvainen2017mean,miyato2018virtual} we establish a benchmark for semi-supervised affordance estimation. 
We synthesize a semi-supervised setting by randomly sampling $1\%$ training data, assumed to be labeled, and the rest are assumed to be unlabeled data. The validation and testing sets are kept the same with standard benchmarks. 

\noindent\textbf{Network and Training.} We utilize DGCNN\cite{wang2019dynamic} as our backbone. During every mini-batch in the training stage, we randomly sample a equal number of labeled data $\matr{X}_{l}$ and unlabeled data $\matr{X}_{ul}$. To fully exploit the unlabeled data, we employ a state-of-the-art semi-supervised learning framework, namely Virtual Adversarial Training (VAT) \cite{miyato2018virtual}. It encourages the consistency between the posterior of unlabeled sample and its augmentation, measured by mean square error,
\vspace{-0.3cm}
\begin{equation}
l_{mse}=\frac{1}{N}\sum_{i}^{M}\sum_{j}^{N}||p_{i,j}-\hat{p}_{i,j}||_2^2
\vspace{-0.2cm}
\end{equation}
where $\hat{p}_{i,j}$ is the posterior prediction for augmented sample. To best exploit the consistency power, the augmentation is obtained by first applying a one step adversarial attack, the corresponding adversarial perturbation is then added to the original point cloud to produce the augmentation. Finally, the total loss for semi-supervised affordance estimation combines both losses defined for labeled data and unlabeled data.
\vspace{-0.1cm}
\begin{equation}
   l = l_{CE} + l_{DICE} + l_{mse}^l + l_{mse}^u
   \vspace{-0.2cm}
\end{equation}
where 
$l_{mse}^l$ and $l_{mse}^u$ is the mean square error (MSE) calculated between labeled and unlabeled data, respectively. We compare the semi-supervised approach against a fully supervised baseline which is trained on the $1\%$ labeled data alone with cross-entropy and dice loss.
We use a mini-batch of 16, 8 for labeled data and 8 for unlabeled data, and follow the same training strategies and hyper-parameters described in~\cite{miyato2018virtual}. We train a full-shape affordance estimation method based on DGCNN only on the labelled data following the description in Sect.~\ref{estimation task}.

\noindent\textbf{Evaluation and Results} We evaluated the methods following the metrics described in Sect.~\ref{estimation task} with results reported in Tab.~\ref{Semi results}. Comparing the performance of semi-supervised affordance estimation to the full-shape one, we found that semi-supervised affordance estimation outperforms the fully supervised baseline on all three metrics. Specifically, the gains for some affordance categories (\eg open) that have low metrics on full-shape affordance estimation are high, which indicates that unlabeled data can provide useful information for affordance learning. In conclusion, we believe that exploiting unlabeled data to improve the performance has practical value and should receive more attention in the future.




\section{Conclusion}

In this work, we proposed 3D AffordanceNet, a 3D point cloud benchmark consisting of 22949 shapes from 23 semantic object categories, annotated with 56307 affordance annotations and covering 18 visual affordance categories. Based on this dataset, we define three individual affordance estimation tasks and benchmarked three state-of-the-art point cloud deep learning networks. The results suggested future research is required to achieve better performance on difficult affordance categories and under SO(3) rotation. Furthermore, we proposed a semi-supervised affordance estimation method to take advantage of large amount of unlabeled data. The proposed dataset encourage the community to focus on affordance estimation research.

\noindent\textbf{Acknowledgement} This work was supported in part by the National Natural Science Foundation of China (Grant No.: 61771201, 61902131), the Program for Guangdong Introducing Innovative and Entrepreneurial Teams (Grant No.: 2017ZT07X183), and the Guangdong R\&D key project of China (Grant No.: 2019B010155001). Xun Xu acknowledges the A*STAR Career Development Award (CDA) Funding for providing financial support (Grant No. 202D8243).
{\small
\bibliographystyle{ieee_fullname}
\bibliography{egbib}

\begin{thebibliography}{10}\itemsep=-1pt

\bibitem{chang2015shapenet}
Angel~X Chang, Thomas Funkhouser, Leonidas Guibas, Pat Hanrahan, Qixing Huang,
  Zimo Li, Silvio Savarese, Manolis Savva, Shuran Song, Hao Su, et~al.
\newblock Shapenet: An information-rich 3d model repository.
\newblock {\em arXiv preprint arXiv:1512.03012}, 2015.

\bibitem{chuang2018learning}
Ching-Yao Chuang, Jiaman Li, Antonio Torralba, and Sanja Fidler.
\newblock Learning to act properly: Predicting and explaining affordances from
  images.
\newblock In {\em Proceedings of the IEEE Conference on Computer Vision and
  Pattern Recognition}, 2018.

\bibitem{do2018affordancenet}
Thanh-Toan Do, Anh Nguyen, and Ian Reid.
\newblock Affordancenet: An end-to-end deep learning approach for object
  affordance detection.
\newblock In {\em IEEE International Conference on Robotics and Automation},
  2018.

\bibitem{earley1970efficient}
Jay Earley.
\newblock An efficient context-free parsing algorithm.
\newblock {\em Communications of the ACM}, 1970.

\bibitem{esteves2018learning}
Carlos Esteves, Christine Allen-Blanchette, Ameesh Makadia, and Kostas
  Daniilidis.
\newblock Learning so (3) equivariant representations with spherical cnns.
\newblock In {\em Proceedings of the European Conference on Computer Vision},
  2018.

\bibitem{gibson1979ecological}
J Gibson~James.
\newblock The ecological approach to visual perception, 1979.

\bibitem{grabner2011makes}
Helmut Grabner, Juergen Gall, and Luc Van~Gool.
\newblock What makes a chair a chair?
\newblock In {\em Proceedings of the IEEE Conference on Computer Vision and
  Pattern Recognition}, 2011.

\bibitem{hassanin2018visual}
Mohammed Hassanin, Salman Khan, and Murat Tahtali.
\newblock Visual affordance and function understanding: A survey.
\newblock {\em arXiv preprint arXiv:1807.06775}, 2018.

\bibitem{hu2016learning}
Ruizhen Hu, Oliver van Kaick, Bojian Wu, Hui Huang, Ariel Shamir, and Hao
  Zhang.
\newblock Learning how objects function via co-analysis of interactions.
\newblock {\em ACM Transactions on Graphics}, 2016.

\bibitem{jain2016structural}
Ashesh Jain, Amir~R Zamir, Silvio Savarese, and Ashutosh Saxena.
\newblock Structural-rnn: Deep learning on spatio-temporal graphs.
\newblock In {\em Proceedings of the IEEE Conference on Computer Vision and
  Pattern Recognition}, 2016.

\bibitem{katz2007direct}
Sagi Katz, Ayellet Tal, and Ronen Basri.
\newblock Direct visibility of point sets.
\newblock In {\em ACM Special Interest Group on GRAPHics and Interactive
  Techniques 2007 papers}. 2007.

\bibitem{koppula2013learning}
Hema~Swetha Koppula, Rudhir Gupta, and Ashutosh Saxena.
\newblock Learning human activities and object affordances from rgb-d videos.
\newblock {\em The International Journal of Robotics Research}, 2013.

\bibitem{koppula2015anticipating}
Hema~S Koppula and Ashutosh Saxena.
\newblock Anticipating human activities using object affordances for reactive
  robotic response.
\newblock {\em IEEE Transactions on Pattern Analysis and Machine Intelligence},
  2015.

\bibitem{laine2016temporal}
Samuli Laine and Timo Aila.
\newblock Temporal ensembling for semi-supervised learning.
\newblock {\em International Conference on Learning Representations}, 2017.

\bibitem{milletari2016v}
Fausto Milletari, Nassir Navab, and Seyed-Ahmad Ahmadi.
\newblock V-net: Fully convolutional neural networks for volumetric medical
  image segmentation.
\newblock In {\em 2016 fourth international conference on 3D vision}, 2016.

\bibitem{miyato2018virtual}
Takeru Miyato, Shin-ichi Maeda, Masanori Koyama, and Shin Ishii.
\newblock Virtual adversarial training: a regularization method for supervised
  and semi-supervised learning.
\newblock {\em IEEE Transactions on Pattern Analysis and Machine Intelligence},
  2018.

\bibitem{mo2019partnet}
Kaichun Mo, Shilin Zhu, Angel~X Chang, Li Yi, Subarna Tripathi, Leonidas~J
  Guibas, and Hao Su.
\newblock Partnet: A large-scale benchmark for fine-grained and hierarchical
  part-level 3d object understanding.
\newblock In {\em Proceedings of the IEEE Conference on Computer Vision and
  Pattern Recognition}, 2019.

\bibitem{myers2015affordance}
Austin Myers, Ching~L Teo, Cornelia Ferm{\"u}ller, and Yiannis Aloimonos.
\newblock Affordance detection of tool parts from geometric features.
\newblock In {\em IEEE International Conference on Robotics and Automation},
  2015.

\bibitem{nguyen2017object}
Anh Nguyen, Dimitrios Kanoulas, Darwin~G Caldwell, and Nikos~G Tsagarakis.
\newblock Object-based affordances detection with convolutional neural networks
  and dense conditional random fields.
\newblock In {\em IEEE/RSJ International Conference on Intelligent Robots and
  Systems}, 2017.

\bibitem{qi2017pointnet++}
Charles~Ruizhongtai Qi, Li Yi, Hao Su, and Leonidas~J Guibas.
\newblock Pointnet++: Deep hierarchical feature learning on point sets in a
  metric space.
\newblock In {\em Advances in neural information processing systems}, 2017.

\bibitem{qi2017predicting}
Siyuan Qi, Siyuan Huang, Ping Wei, and Song-Chun Zhu.
\newblock Predicting human activities using stochastic grammar.
\newblock In {\em Proceedings of the IEEE International Conference on Computer
  Vision}, 2017.

\bibitem{roy2016multi}
Anirban Roy and Sinisa Todorovic.
\newblock A multi-scale cnn for affordance segmentation in rgb images.
\newblock In {\em Proceedings of the European Conference on Computer Vision},
  2016.

\bibitem{sawatzky2017weakly}
Johann Sawatzky, Abhilash Srikantha, and Juergen Gall.
\newblock Weakly supervised affordance detection.
\newblock In {\em Proceedings of the IEEE Conference on Computer Vision and
  Pattern Recognition}, 2017.

\bibitem{song2015learning}
Hyun~Oh Song, Mario Fritz, Daniel Goehring, and Trevor Darrell.
\newblock Learning to detect visual grasp affordance.
\newblock {\em IEEE Transactions on Automation Science and Engineering}, 2015.

\bibitem{tarvainen2017mean}
Antti Tarvainen and Harri Valpola.
\newblock Mean teachers are better role models: Weight-averaged consistency
  targets improve semi-supervised deep learning results.
\newblock In {\em Advances in neural information processing systems}, 2017.

\bibitem{vu2014predicting}
Tuan-Hung Vu, Catherine Olsson, Ivan Laptev, Aude Oliva, and Josef Sivic.
\newblock Predicting actions from static scenes.
\newblock In {\em Proceedings of the European Conference on Computer Vision},
  2014.

\bibitem{wang2019dynamic}
Yue Wang, Yongbin Sun, Ziwei Liu, Sanjay~E Sarma, Michael~M Bronstein, and
  Justin~M Solomon.
\newblock Dynamic graph cnn for learning on point clouds.
\newblock {\em ACM Transactions on Graphics}, 2019.

\bibitem{xie2020pointcontrast}
Saining Xie, Jiatao Gu, Demi Guo, Charles~R Qi, Leonidas Guibas, and Or Litany.
\newblock Pointcontrast: Unsupervised pre-training for 3d point cloud
  understanding.
\newblock In {\em Proceedings of the European Conference on Computer Vision},
  2020.

\bibitem{yi2016scalable}
Li Yi, Vladimir~G Kim, Duygu Ceylan, I-Chao Shen, Mengyan Yan, Hao Su, Cewu Lu,
  Qixing Huang, Alla Sheffer, and Leonidas Guibas.
\newblock A scalable active framework for region annotation in 3d shape
  collections.
\newblock {\em ACM Transactions on Graphics}, 2016.

\bibitem{Yifan_2019_CVPR}
Wang Yifan, Shihao Wu, Hui Huang, Daniel Cohen-Or, and Olga Sorkine-Hornung.
\newblock Patch-based progressive 3d point set upsampling.
\newblock In {\em Proceedings of the IEEE Conference on Computer Vision and
  Pattern Recognition}, 2019.

\bibitem{yu2019partnet}
Fenggen Yu, Kun Liu, Yan Zhang, Chenyang Zhu, and Kai Xu.
\newblock Partnet: A recursive part decomposition network for fine-grained and
  hierarchical shape segmentation.
\newblock In {\em Proceedings of the IEEE Conference on Computer Vision and
  Pattern Recognition}, 2019.

\bibitem{zhou2019semantic}
Bolei Zhou, Hang Zhao, Xavier Puig, Tete Xiao, Sanja Fidler, Adela Barriuso,
  and Antonio Torralba.
\newblock Semantic understanding of scenes through the ade20k dataset.
\newblock {\em International Journal of Computer Vision}, 2019.

\bibitem{zhu2015understanding}
Yixin Zhu, Yibiao Zhao, and Song Chun~Zhu.
\newblock Understanding tools: Task-oriented object modeling, learning and
  recognition.
\newblock In {\em Proceedings of the IEEE Conference on Computer Vision and
  Pattern Recognition}, 2015.

\end{thebibliography}
}

\clearpage
\textbf{\Large Appendix} \\
\appendix

This supplementary material provides additional dataset visualization, qualitative results, technical details, full question list and the interface of annotation tool.

In Sec.~\ref{complete list} we present all questions visible to annotators during data annotation procedure. Data annotation agreement is discuss in Sec.\ref{sec:annotation}. We further provide more ground-truth visualizations for each shape category in 3D AffordanceNet dataset in Sec.~\ref{dataset visualization}. The Sec.~\ref{qualitative examples} presents more qualitative examples of full-shape, partial-view and rotation-invariant affordance estimation results with PointNet++ and DGCNN as backbones. In Sec.~\ref{training details} we describe more details about neural network architecture and training parameters.  We visualize the annotation interface and the main components of our web-based annotation tool in Sec.~\ref{gui interface}. Then we explore the performance gain benefit from fine-tuning in Sec.~\ref{pc finetune}. At last we demonstrate why a truly functional understanding of object affordance requires learning and prediction in the 3D physical domain in Sec.~\ref{understanding in 3d}.

\section{Complete List of Questions} \label{complete list}

We list all questions that the annotation interface displays to annotators. The complete question list is shown in Tab.~\ref{complete question list}. The questions that the annotation interface proposes directly determine how the annotators understand the affordance, thus we carefully define the questions for each affordance. 

\begin{table*}[t]
\centering
\resizebox{\textwidth}{!}{%
    \begin{tabular}{c|c|c}
    \hline
    \textbf{Object}                   & \textbf{Affordance} & \textbf{Question}                                                                                                                                \\ \hline
    \multirow{3}{*}{Bed}              & Lay                 & If you were to lie on this bed, which points would you lie on the bed?                                                                           \\
                                      & Sit                 & If you were to sit on this bed, at which points on the bed would you sit?                                                                        \\
                                      & Support             & If you put something on this bed, at which points on the bed would you put it?                                                                   \\ \hline
    \multirow{4}{*}{Bag}              & Grasp               & If you want to grab the bag, at which points will your palm position be?                                                                         \\
                                      & Lift                & If you want to lift the bag, at which point is your finger most likely to carry the bag?                                                         \\
                                      & Contain             & If you package things into the bag, at which points in this bag would you put?                                                                   \\
                                      & Open                & If you want to open the bag, from which points of the package would you open it?                                                                 \\ \hline
    \multirow{5}{*}{Bottle}           & Grasp               & If you grab the bottle, at which points on the bottle handle will your palm touch?                                                               \\
                                      & Wrap-Grasp          & If you wrap-grasp the bottle, at which points on the bottle wall will your palm touch?                                                           \\
                                      & Contain             & If you pour water into the bottle, which points will the water first touch when it falls into the bottle?                                        \\
                                      & Open                & If you want to open this bottle, from which points on the cap would you open it?                                                                 \\
                                      & Pour                & Suppose there is water in the bottle, and you want to pour the water out of the bottle. From which points on the bottle will the water flow out? \\ \hline
    \multirow{3}{*}{Bowl}             & Wrap-Grasp          & If you wrap-grasp the bowl, at which points on the bowl wall will your palm touch?                                                               \\
                                      & Contain             & If you want to put something in the bowl, at which points in the bowl would you put it?                                                          \\
                                      & Pour                & Suppose there is water in the bowl, and you want to pour the water out of the bowl. From which points on the bowl will the water flow out?       \\ \hline
    \multirow{3}{*}{Chair}            & Sit                 & If you were sitting on this chair, on which points would you sit?                                                                                \\
                                      & Support             & If you want to put something on the chair, at which points on the chair would you put it?                                                        \\
                                      & Move                & If you want to move this chair, at which points on the chair will you exert force?                                                               \\ \hline
    Clock                             & Display             & If you want to look at the time, which points on this clock would you look at?                                                                   \\ \hline
    \multirow{2}{*}{Dishwasher}       & Contain             & If you want to load things in the dishwasher, at which points in the dishwasher would you put the things?                                        \\
                                      & Open                & If you want to open this dishwasher, from which points on the dishwasher door would you open it?                                                 \\ \hline
    Display                           & Display             & If you look on the screen, which points on the screen will you look at?                                                                          \\ \hline
    \multirow{3}{*}{Door}             & Push                & If you want to push the door, at which points on the door will your palm touch?                                                                  \\
                                      & Open                & If you were to open the door, from which points on the door would you open it?                                                                   \\
                                      & Pull                & If you want to pull the door, which points on the door will you pull with your finger?                                                           \\ \hline
    \multirow{2}{*}{Earphone}         & Grasp               & If you want to grab this earphone, where will your palm position be?                                                                             \\
                                      & Listen              & If you want to listen to music with headphones, which points on the headphones will point to your ears?                                          \\ \hline
    \multirow{2}{*}{Faucet}           & Grasp               & If you want to grab this faucet, which points on the faucet will your palm touch?                                                                \\
                                      & Open                & If you want to boil water, at which points on the tap would you open the water valve?                                                            \\ \hline
    \multirow{2}{*}{Hat}              & Grasp               & If you want to grab this hat, which points on the hat will your palm touch?                                                                      \\
                                      & Wear                & If you want to wear this hat, which points on the hat will make contact with your head?                                                          \\ \hline
    Keyboard                          & Press               & If you want to press keys on the keyboard, which points on the keyboard would you press?                                                         \\ \hline
    \multirow{3}{*}{Knife}            & Grasp               & If you want to grab this knife, at which points on the handle will your palm touch?                                                              \\
                                      & Cut                 & If you want to cut something with this knife, which points on the blade will come into contact with the object?                                  \\
                                      & Stab                & If you use this knife to poke an object, which points on the blade will come into contact with the object?                                       \\ \hline
    \multirow{2}{*}{Laptop}           & Display             & If you look on the computer screen, which points on the screen will you look at?                                                                 \\
                                      & Press               & If you want to press keys on a computer keyboard, which points on the keyboard would you press?                                                  \\ \hline
    \multirow{2}{*}{Microwave}        & Open                & If you want to open the microwave, from which points on the microwave door would you open it?                                                    \\
                                      & Contain             & If you put something in the microwave, at which points in the microwave would you put the object?                                                \\ \hline
    \multirow{4}{*}{Mug}              & Pour                & Suppose there is water in the mug, and you want to pour the water out of the mug. From which points on the mug will the water flow?              \\
                                      & Contain             & If you pour water into the mug, which points will the water first touch when it falls into the mug?                                              \\
                                      & Wrap-Grasp          & If you wrap-grasp this mug with your hand, which points on the mug will your palm touch?                                                         \\
                                      & Grasp               & If you grab this mug, which points on the mug handle will your palm touch?                                                                       \\ \hline
    \multirow{2}{*}{Refrigerator}     & Contain             & If you put things in the refrigerator, at which points in the refrigerator would you put?                                                        \\
                                      & Open                & If you want to open the refrigerator, from which points on the refrigerator door would you open it?                                              \\ \hline
    \multirow{3}{*}{Scissors}         & Grasp               & If you want to grab this scissors, which points on the handle of the scissors will your palm touch?                                              \\
                                      & Cut                 & If you want to use scissors to cut something, which points on the scissors blade will contact the object?                                        \\
                                      & Stab                & If you poke an object with this pair of scissors, which points on the blade will come into contact with the object?                              \\ \hline
    \multirow{2}{*}{StorageFurniture} & Contain             & If you want to put something in the cabinet, at which points in the cabinet would you put it?                                                    \\
                                      & Open                & If you want to open this cabinet, from which points on the cabinet door would you open it?                                                       \\ \hline
    \multirow{2}{*}{Table}            & Support             & If you want to put something on the table, at which points on the table would you put the object?                                                \\
                                      & Move                & If you want to move this table, at which points on this table will you exert your strength                                                       \\ \hline
    \multirow{3}{*}{TrashCan}         & Contain             & If you put trash in the trash can, which points will the trash drop first touch?                                                                 \\
                                      & Pour                & If you want to dump out the trash in the trash can, at which points on the trash can will the trash slip out?                                    \\
                                      & Open                & If you want to open the lid of this trash can, from which points on the trash can you open it?                                                   \\ \hline
    \multirow{3}{*}{Vase}             & Contain             & If you pour water into the vase, which points will the water first touch when it falls into the vase?                                            \\
                                      & Pour                & Suppose there is water in the vase, and you want to pour the water out of the vase. From which points on the vase will the water flow out?       \\
                                      & Wrap-Grasp          & If you wrap-grasp this vase with your hands, which points on the vase will your palm touch?                                                      \\ \hline
    \end{tabular}}
    \caption{The complete list of questions that the annotation interface proposes to annotators}
    \label{complete question list}
\end{table*}

\section{Data Annotation}\label{sec:annotation}
\begin{table}[!htb]
\resizebox{\linewidth}{!}{%
\begin{tabular}{c|ccccccccc}
\hline
\textbf{Affordance} & \textbf{Grasp}   & \textbf{Lift} & \textbf{Contain} & \textbf{Open}   & \textbf{Lay}  & \textbf{Sit}   & \textbf{Support} & \textbf{Wrap.} & \textbf{Pour} \\ \hline
\textbf{IOU}        & 22.7             & 35.9          & 13.8             & 28.3            & 26.4          & 18.8           & 14.4             & 6.8            & 14.2          \\
\textbf{Variance}   & 0.007            & 0.004         & 0.002            & 0.003           & 0.005         & 0.004          & 0.005            & 0.005          & 0.004         \\ \hline
\textbf{Affordance} & \textbf{Display} & \textbf{Push} & \textbf{Pull}    & \textbf{Listen} & \textbf{Wear} & \textbf{Press} & \textbf{Move}    & \textbf{Cut}   & \textbf{Stab} \\ \hline
\textbf{IOU}        & 39.8             & 13.3          & 34.1             & 17.4            & 7.3           & 29.7           & 12.3             & 16             & 3.5           \\
\textbf{Variance}   & 0.005            & 0.005         & 0.002            & 0.002           & 0.006         & 0.18           & 0.009            & 0.004          & 0.001         \\ \hline
\end{tabular}%
}
\caption{IOU between each affordance and a best combination of parts. Variance of each affordance category. Numbers of IOU are in $\%$.}
\label{IOU_Variance}
\vspace{-0.4cm}
\end{table}
We demonstrate that the human perceived affordance often do not fully overlap with the individual part specified in PartNet dataset. To investigate the relation between parts and affordances, we report the maximal IoU between each affordance and a best combination of parts (most fine-grained level) in Tab.\ref{IOU_Variance}. The low IoU indicates that it is impossible to combine several parts to derive affordance, suggesting the necessity to label affordance from scratch.

{We have 3 annotators to label each object, which gives certain diversities and partially addresses the issue of ambiguities. We report the inconsistency as average variance across 3 annotators in Tab.\ref{IOU_Variance}. The relatively smaller variance of each affordance indicates that there exists patterns in the annotations, and models can therefore learn affordance from data.}

\section{More Ground-Truth Visualizations} \label{dataset visualization}

\begin{figure*}[t]
   \begin{center}
      \includegraphics[width=0.8\textwidth]{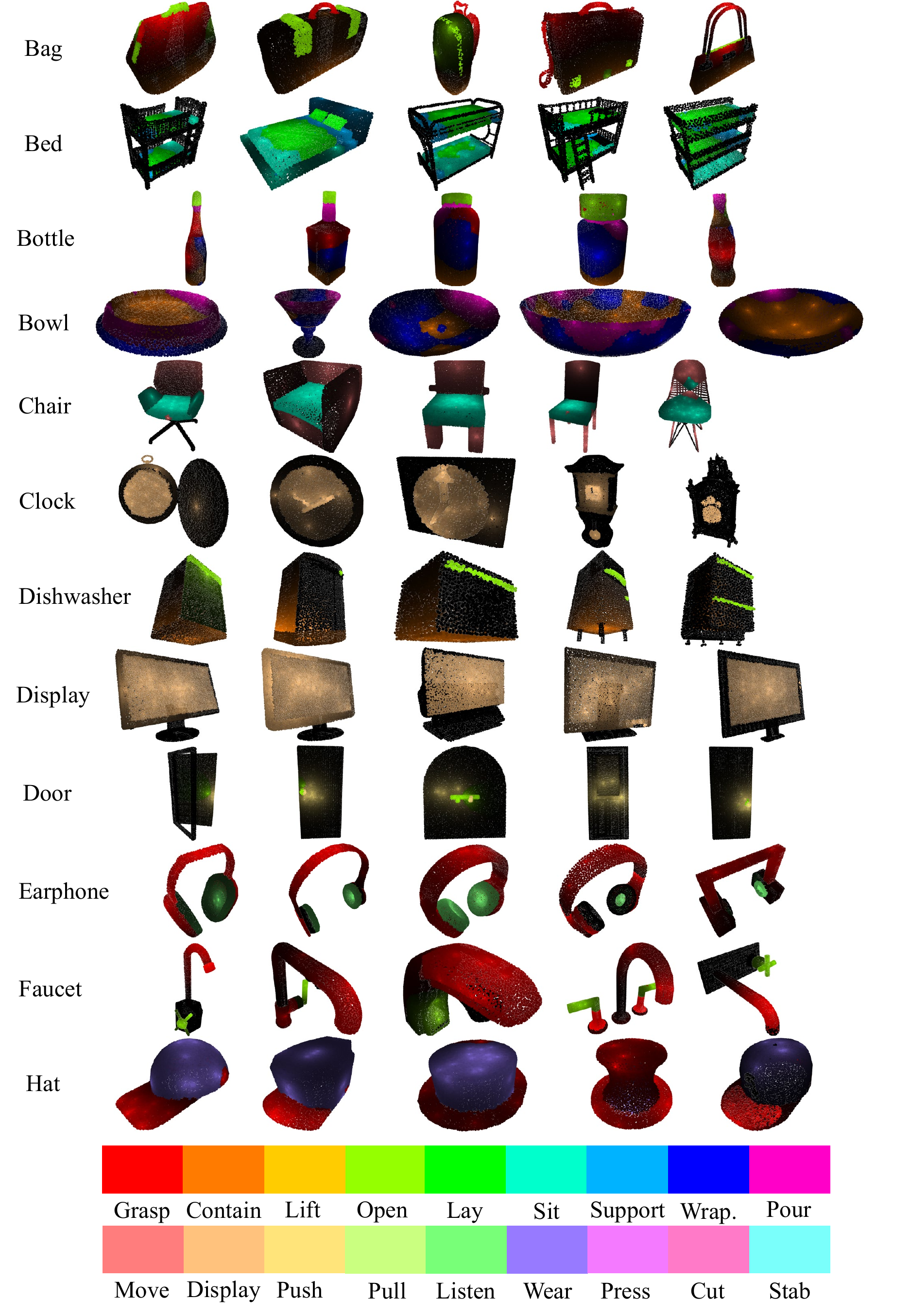}
   \end{center}
   \caption{Ground Truth data visualization(1/2).}
   \label{ground truth 1}
\end{figure*}

\begin{figure*}[t]
   \begin{center}
      \includegraphics[width=0.8\textwidth]{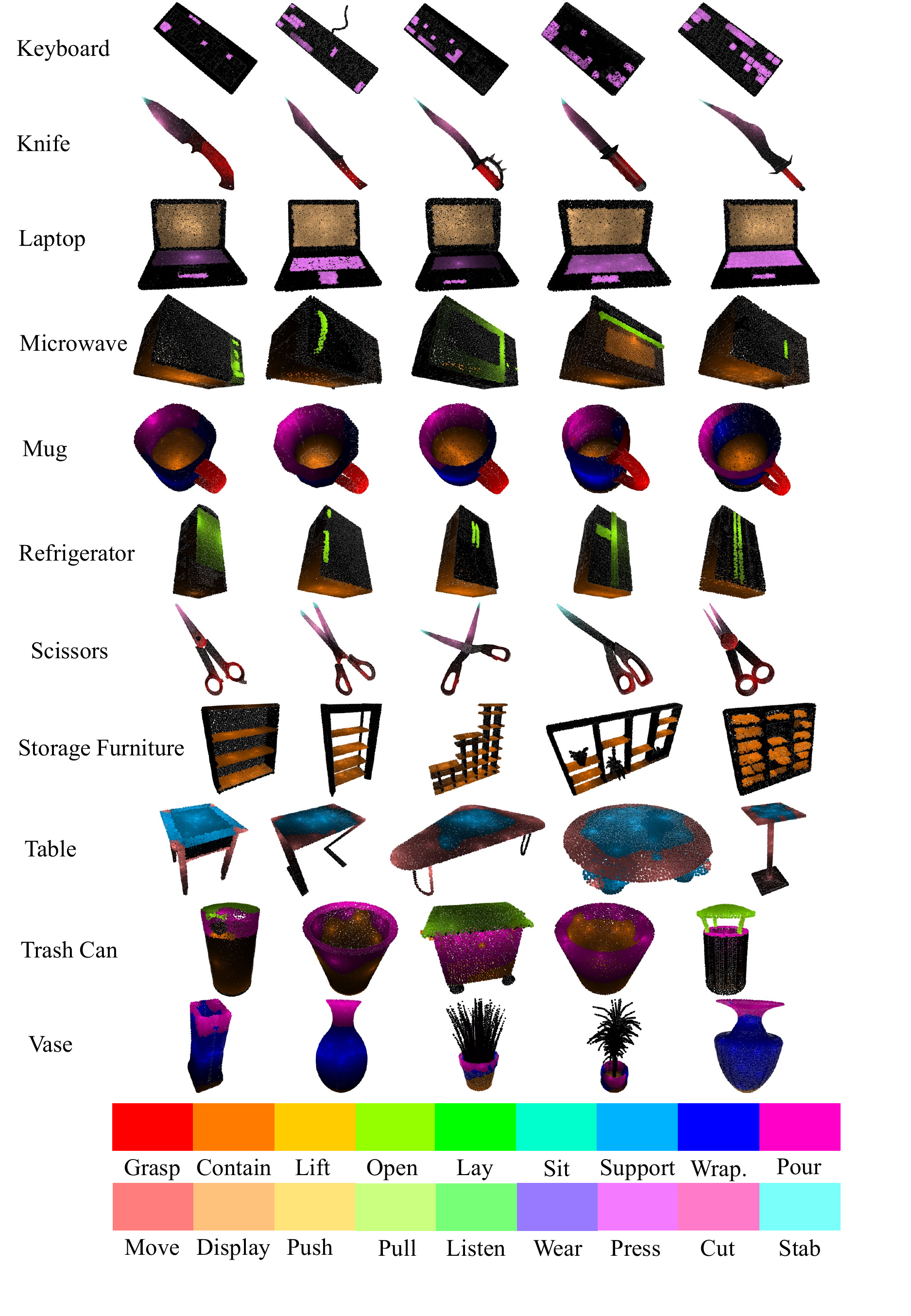}
   \end{center}
   \caption{Ground Truth data visualization(2/2).}
   \label{ground truth 2}
\end{figure*}

We present more ground-truth visualization in Fig.~\ref{ground truth 1} and Fig.~\ref{ground truth 2}. From the visualization of ground-truth, we can observe that the human perceived affordances often do not fully overlap with the individual parts specified in PartNet dataset, therefore it justifies the need to annotate affordance separately from existing part annotations.

\begin{figure*}[t]
   \begin{center}
      \includegraphics[width=1.0\textwidth]{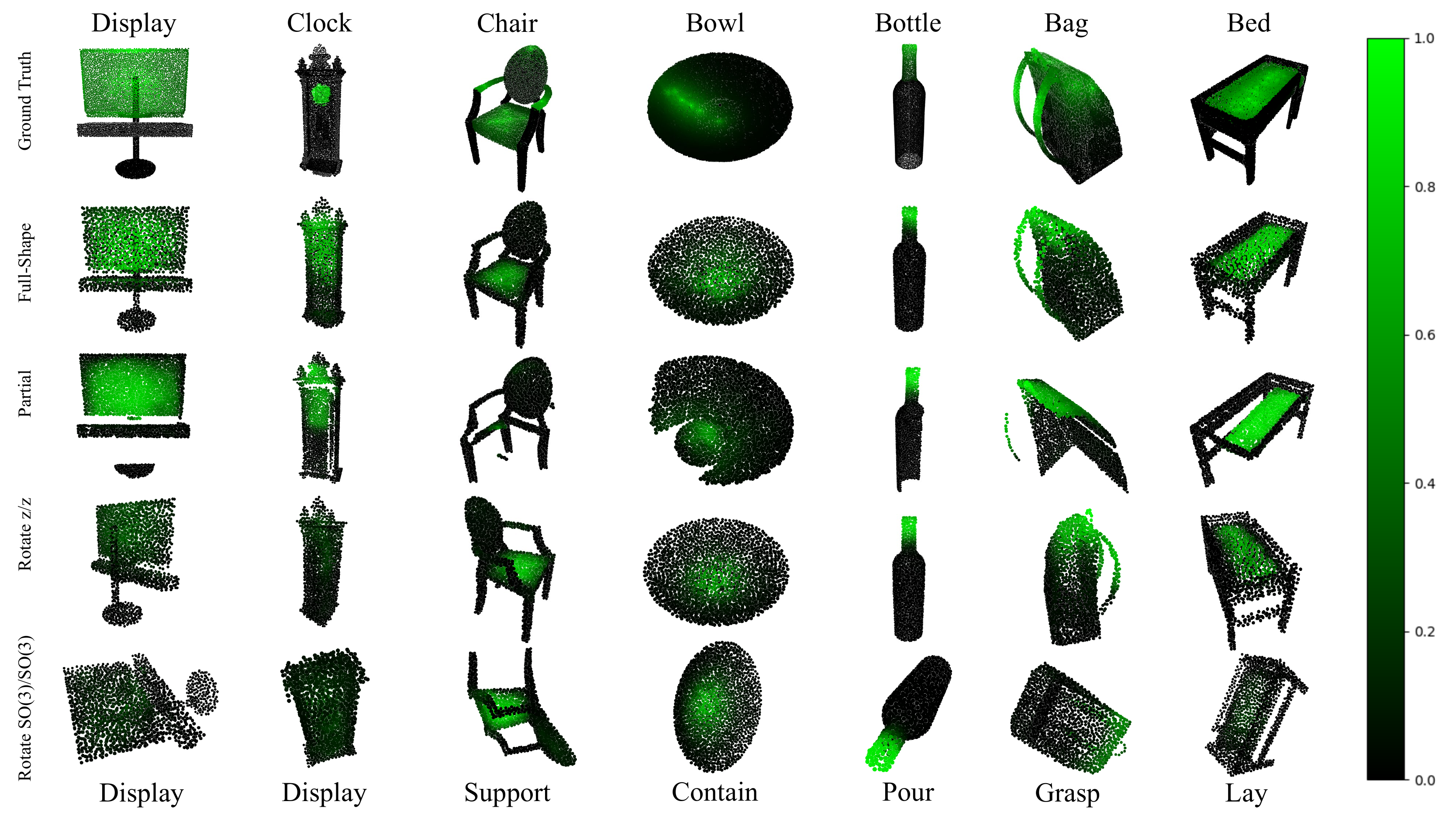}
   \end{center}
   \caption{Qualitative results for affordance estimation from PointNet++(1/2). The top row shows the ground-truth. The second row shows the full-shape estimated results, the third row shows the partial-view estimated result, the fourth and the bottom row show the $z/z$ and $SO(3)/SO(3)$ rotation-invariant estimated results, respectively. All results come from PointNet++. The top words indicate the semantic category of each column and the bottom words indicate the affordance category. The greener the color of the points, the higher the confidence about specific affordance types. \textit{Wrap.} is the abbreviation of \textit{Wrap-Grasp.}}
   \label{estimation image 1}
\end{figure*}

\begin{figure*}[t]
   \begin{center}
      \includegraphics[width=0.93\textwidth]{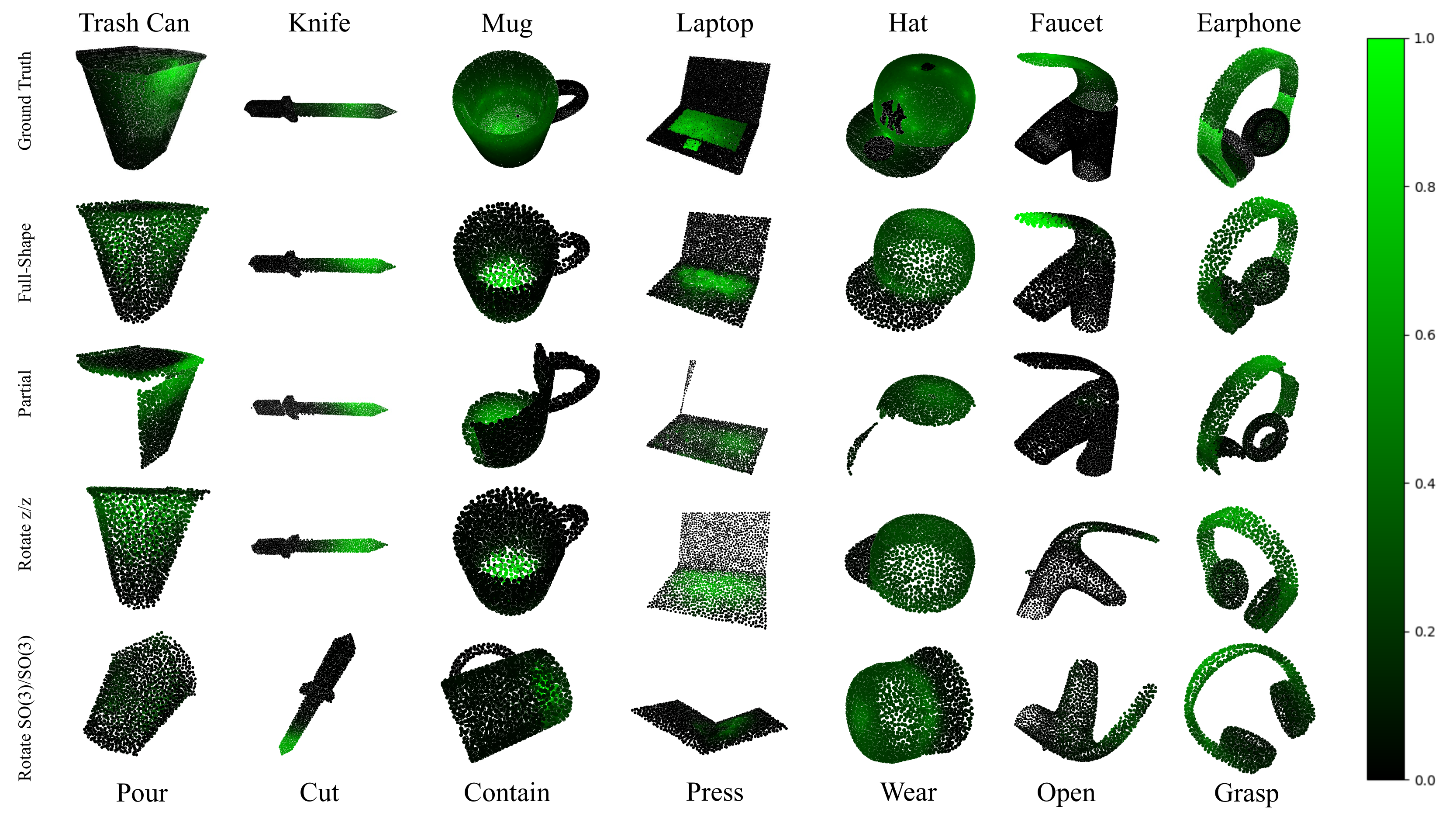}
   \end{center}
   \vspace{-0.5cm}
   \caption{Qualitative results for affordance estimation from PointNet++(2/2). The top row shows the ground truth. The second row shows the full-shape estimated results, the third row shows the partial-view estimated result, the fourth and the bottom row show the $z/z$ and $SO(3)/SO(3)$ rotation-invariant estimated results, respectively. All results come from PointNet++. The top words indicate the semantic category of each column and the bottom words indicate the affordance category. The greener the color of the points, the higher the confidence about specific affordance types. \textit{Wrap.} is the abbreviation of \textit{Wrap-Grasp.}}
   \label{estimation image 2}
\end{figure*}

\begin{figure*}[t]
   \begin{center}
      \includegraphics[width=0.93\textwidth]{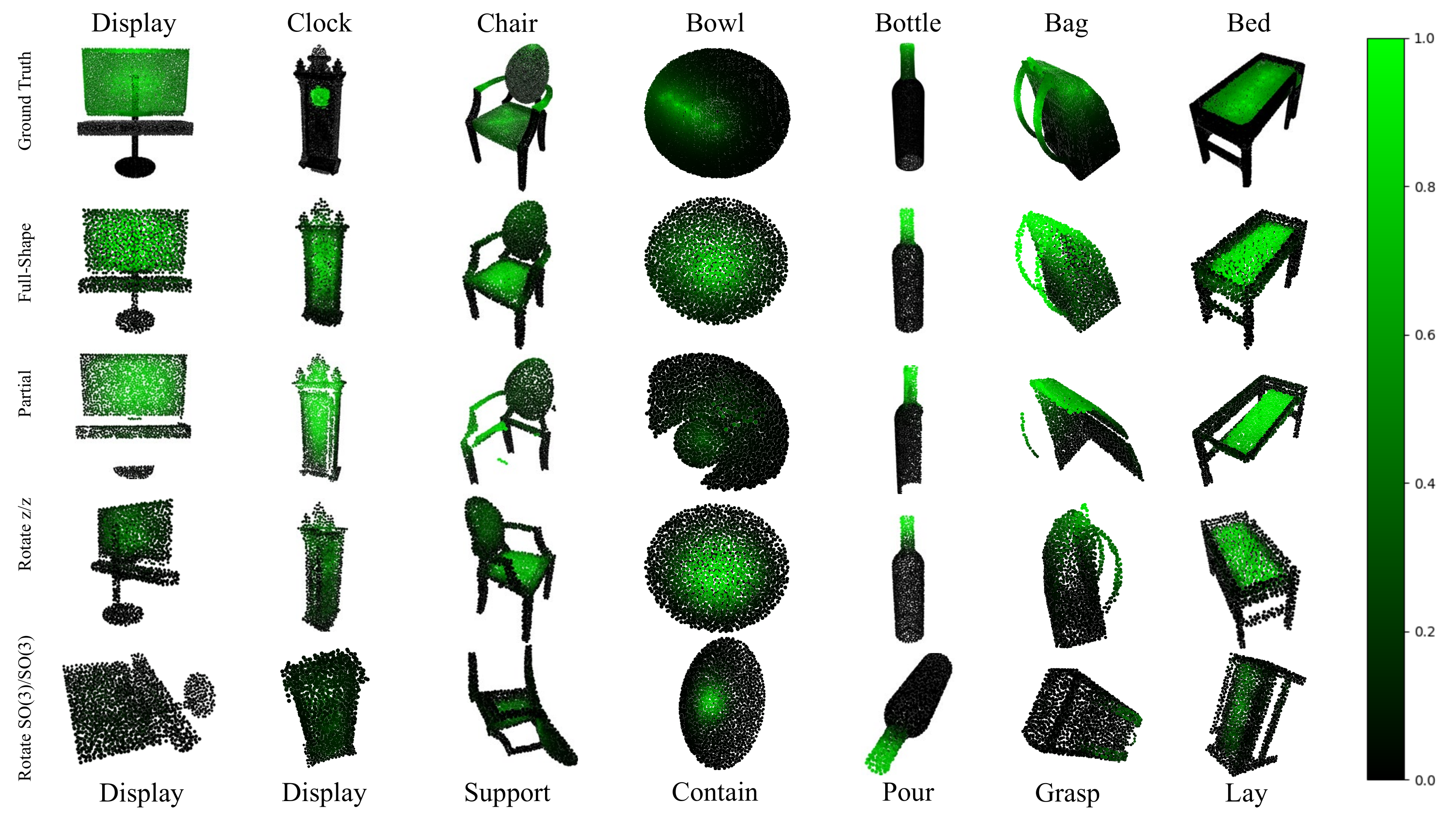}
   \end{center}
   \vspace{-0.5cm}
   \caption{Qualitative results for affordance estimation from DGCNN(1/2). The top row shows the ground truth. The second row shows the full-shape estimated results, the third row shows the partial-view estimated result, the fourth and the bottom row show the $z/z$ and $SO(3)/SO(3)$ rotation-invariant estimated results, respectively. All results come from DGCNN. The top words indicate the semantic category of each column and the bottom words indicate the affordance category. The greener the color of the points, the higher the confidence about specific affordance types. \textit{Wrap.} is the abbreviation of \textit{Wrap-Grasp.}}
   \label{estimation image 3}
\end{figure*}

\begin{figure*}[t]
   \begin{center}
      \includegraphics[width=1.0\textwidth]{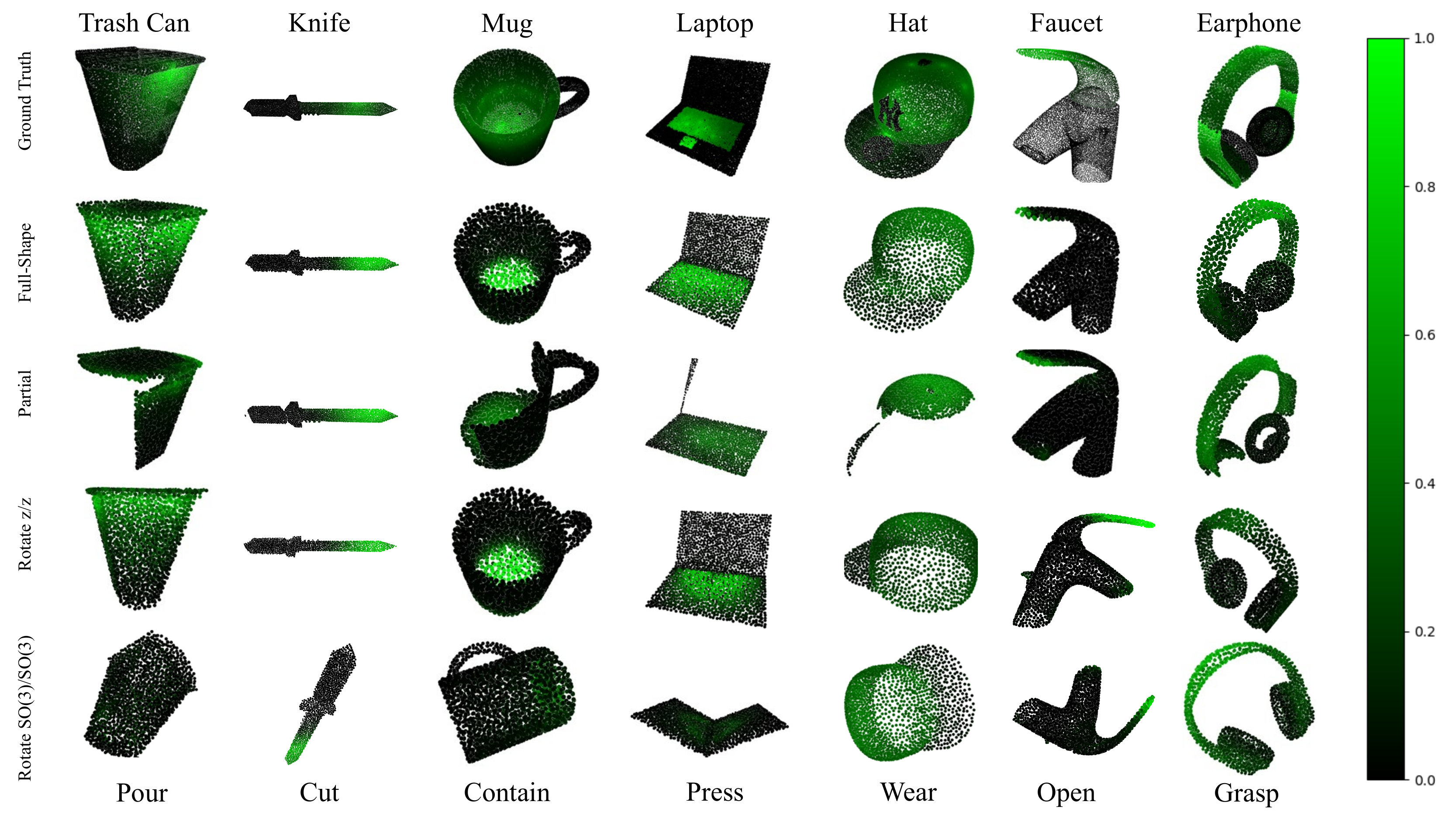}
   \end{center}
   \caption{Qualitative results for affordance estimation from DGCNN(2/2). The top row shows the ground truth. The second row shows the full-shape estimated results, the third row shows the partial-view estimated result, the fourth and the bottom row show the $z/z$ and $SO(3)/SO(3)$ rotation-invariant estimated results, respectively. All results come from DGCNN. The top words indicate the semantic category of each column and the bottom words indicate the affordance category. The greener the color of the points, the higher the confidence about specific affordance types. \textit{Wrap.} is the abbreviation of \textit{Wrap-Grasp.}}
   \label{estimation image 4}
\end{figure*}

\begin{figure*}[t]
   \begin{center}
      \includegraphics[width=1.0\textwidth]{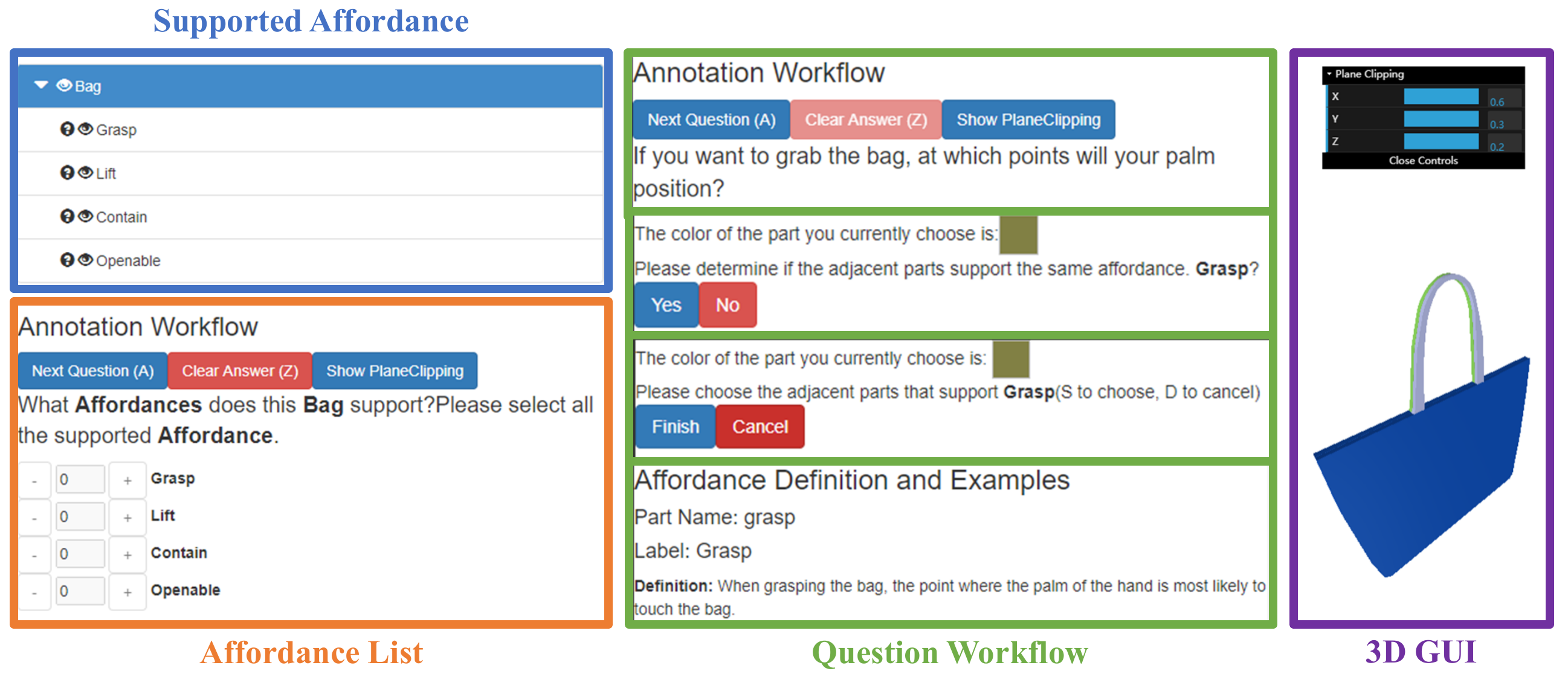}
   \end{center}
   \caption{The annotation interface of our web-based annotation tool. We show the GUI and main component of the annotation interface.}
   \label{annotation interface}
\end{figure*}

\section{More Qualitative Examples} \label{qualitative examples}
We present more qualitative examples for full-shape, partial-view and rotation-invariant affordance estimation experiments with both PointNet++ and DGCNN as backbone in Fig.~\ref{estimation image 1},~\ref{estimation image 2},~\ref{estimation image 3} and ~\ref{estimation image 4}.

The estimation results from PointNet++ and DGCNN are quite interesting. The predicted affordance locations from the two networks are close while the confidences of the points belonging to specific affordances have different tendencies. In many cases, \eg \textit{grasp} for \textit{bag} and \textit{press} for \textit{laptop}, PointNet++ tends to predict scores with low confidence which will cause more false-negative predictions while DGCNN predicts scores more aggressively, leading to more false-positive examples. 

\section{Training Details} \label{training details}

In this section we describe more details of training procedure. We conduct all experiments using the segmentation branch of PointNet++ and DGCNN as shared backbones. 

In specific, the dimension of point-wise features by PointNet++ and DGCNN are 128 and 256, respectively. We formulate affordance estimation as a binary classification problem, therefore, we set up classification heads for each affordance category, in total there are 18 classification heads which have the same architecture with different parameters. Specifically, the classification head for each affordance category consists of ${FC(m,128),FC(128,1)}$, where the function $FC(x,y)$ denotes a fully connected layer that takes x dimension vectors as inputs and outputs y dimension vectors, the number $m$ denotes the dimension of point-wise features by the shared backbones (in our case, it will be 128 for PointNet++ and 256 for DGCNN). In practice, the first $FC$ is followed by a Batch Normalization layer.

For PointNet++, we set the training learning rate 0.001 and optimize the parameters with Adam optimizer, the learning rate is reduced by half every 20 epochs, we train the network for 200 epochs, the batch size is 16. The weight decay for Adam optimizer is set to 1e-8. For DGCNN, we set the learning rate to  0.1 and optimize the parameters with SGD optimizer, the momentum and weight decay for SGD are set as 0.9 and 1e-4, respectively. We use a cosine annealing algorithm to adjust the learning rate where the algorithm can be described as followed:
\begin{equation}
    \eta_{t} = \eta_{min} + \frac{1}{2}(\eta_{max}-\eta_{min})(1+cos(\frac{T_{cur}}{T_{max}}\pi))
\end{equation}
where $\eta_{t}$ is the adjusted learning rate, $\eta_{min}$ is the minimum learning rate, $\eta_{max}$ is set to the initial learning rate, $T_{cur}$ is the number of current epochs. We set the batch size to 16 and train the network for 200 epochs. 

Particularly, for semi-supervised affordance estimation, we use DGCNN as shared backbone and follow the training strategies described above. We set the batch size to 16, 8 for labeled data and 8 for unlabeled data. We set the $\xi$ and $\epsilon$ of Virtual Adversarial Training to 1e-6 and 2.0, which are the default hyper-parameters described in its paper. We calculate the virtual adversarial direction in 1 iteration which is recommended by the original paper. We implement all experiments with PyTorch.

\section{GUI Interface for Annotation} \label{gui interface}

We show the GUI interface of web-based annotation tool in Fig.~\ref{annotation interface}. We manually modify the annotation system released by PartNet fit our requirements. We color the parts of shapes according to the pre-defined colormap in PartNet dataset. 

The annotators can observe the geometric information of shapes and are able to freely translate, rotate and change the scale of shapes in \textbf{3D GUI}. In \textbf{Question Workflow}, the annotation interface asks the annotators some questions to guide them to select keypoints on the surface of shapes. From \textbf{Supported Affordance}, the annotators can check the supported affordances that are determined by selecting the corresponding affordances in \textbf{Affordance List}. 

\section{PointContrast Fine-Tune} \label{pc finetune}

\begin{table}[!htb]
\resizebox{\linewidth}{!}{%
\begin{tabular}{c|cccc|l|cccc}
\hline
          & mAP  & mAUC & aIOU & MSE   &                              & mAP  & mAUC & aIOU & MSE   \\ \hline
Fine-Tune & \textbf{47.4} & \textbf{86.3} & \textbf{19.7} & \textbf{0.063} & \multicolumn{1}{c|}{Scratch} & 45.9 & 85.8 & 19.1 & 0.064 \\ \hline
\end{tabular}%
}
\caption{The full-shape affordance estimation performance comparison between the U-Net fine-tuned on our dataset and the U-Net trained from scratch.}
\label{pointcontrast_compare}
\end{table}

In 2D vision, in order to boost the performance, it is popular to fine-tune a network on the smaller target set where the network was pre-trained on a rich source set. Recently, PointContrast shows that by pre-training the network on the ScanNet dataset using contrastive learning, the pre-trained network can achieve the state-of-the art performances via fine-tuning on several downstream tasks. To explore the opportunity of boosting performances of affordance estimation by fine-tuning the pre-trained network on our 3D AffordanceNet dataset, we utilize the U-Net architecture and the pre-trained weight provided by PointContrast. We then fine-tune the network using SGD optimizer with learning rate=0.1, weight decay=1e-4 and momentum=0.9 for 60 epochs. The loss function that we use to fine-tune the network is the same as the one proposed in the Section 4.1. We also train the network straightly from scratch with the same network architecture for the comparison.

Tab.~\ref{pointcontrast_compare} reports the performances of both fine-tuned and trained-from-scratch U-Net on full-shape affordance estimation task. The results show that the performances of the fine-tuned one surpass the one that is training from scratch, meaning that the network can benefit from the pre-training on a rich source dataset during the fine-tune process on the affordance estimation task, which may also works for the other networks such as PointNet++ and DGCNN.

\section{Affordance Understanding in 3D} \label{understanding in 3d}

\begin{table}[!htb]
\resizebox{\linewidth}{!}{%
\begin{tabular}{c|ccc|c|ccc}
\hline
\textbf{}        & \textbf{mAP} & \textbf{AUC} & \textbf{aIOU} & \textbf{}     & \textbf{mAP} & \textbf{AUC} & \textbf{aIOU} \\ \hline
\textbf{P 3DV} & 48.0         & 87.4           & 19.3            & \textbf{D 3DV} & 46.4         & 85.5         & 17.8            \\
\textbf{P MTV}  & 45.1         & 84.4         & 16.6          & \textbf{D MTV}  & 41.6         & 82.3         & 13.4          \\
\textbf{P SGV}  & 35.0         & 77.8         & 12.9          & \textbf{D SGV}  & 35.0         & 78.8         & 11.5          \\ \hline
\end{tabular}%
}
\caption{The comparisons between 3D and 2.5D. P and D refer to PointNet++ and DGCNN respectively.
}
\label{Multiview}
\vspace{-0.3cm}
\end{table}

Previous works on affordance understanding focus on learning affordances in 2D or 2.5D domain, however, many types of affordance are related to functional attributes of objects, and the relevant actions can only be accomplished given 3D affordance reasoning on the object surface. For example, a successful grasp of mug replies on inference of surface grasp points (i.e., prediction of the grasp affordance) that may be self-occluded in a single-view observation (i.e., 2.5D). Annotated 3D affordance data facilitate reasoning on the complete object surface.

To quantify the benefit, we conduct the following experiments based on our dataset. We randomly sample one single view (2.5D) from each object for training, namely single-view partial (SGV), and randomly sample 4 views from each object, namely multi-view partial (MTV), then we test the SGV/MTV models on full-shape data. We compare SGV and MTV with training on full 3D data (3DV). Results in Tab.~\ref{Multiview} verify our analysis. It is worth noting that the ground-truth of single view (2.5D) also relies on 3D annotation.

\end{document}